\documentclass[10pt,twocolumn,letterpaper]{article}

\usepackage{cvpr}
\usepackage{times}
\usepackage{epsfig}
\usepackage{graphicx}
\usepackage{amsmath}
\usepackage{amssymb}

\usepackage{booktabs}

\usepackage{cite}

\usepackage[pagebackref=true,breaklinks=true,letterpaper=true,colorlinks,bookmarks=false]{hyperref}

\cvprfinalcopy 

\def\cvprPaperID{1881} 

\ifcvprfinal\fi
\begin{document}

\title{HOPE-Net: A Graph-based Model for Hand-Object Pose Estimation}

\author{Bardia Doosti\(^1\)  \and 
Shujon Naha\(^1\)   \and 
Majid Mirbagheri\(^2\)  \and 
David J. Crandall\(^1\)  \and 
\(^1\) Luddy School of Informatics, Computing, and Engineering, Indiana University Bloomington\\
{\tt\small \{bdoosti,snaha,djcran\}@indiana.edu}\\
\(^2\) Institute for Learning and Brain Sciences, University of Washington\\
{\tt\small mbagheri@uw.edu}\\
\small Project Page: \url{http://vision.sice.indiana.edu/projects/hopenet}
}

\maketitle

\begin{abstract}
Hand-object pose estimation (HOPE) aims to jointly detect the poses of both
a hand and of a held object.  In this paper, we
propose a lightweight model called \textit{HOPE-Net} which
jointly estimates hand and object pose in 2D and 3D in
real-time. Our network uses a cascade of two adaptive graph
convolutional neural networks, one to estimate 2D coordinates of the
hand joints and object corners, followed by another to convert 2D
coordinates to 3D. Our experiments show that through end-to-end training of
the full network, we achieve better accuracy for both the 2D and
3D coordinate estimation problems. The proposed 2D to 3D graph convolution-based 
model could be applied to other 3D landmark detection problems, where
it is possible to first predict the 2D keypoints and then transform
them to 3D. 
\end{abstract}

\section{Introduction}

We use our hands as a primary means of sensing and interacting
with the world.  Thus to understand human activity, computer vision systems
need to be able to detect the pose of the hands and to identify
properties of the objects that are being handled. This human Hand-Object Pose Estimation (HOPE) problem is
crucial for a variety of applications, including augmented and virtual
reality, fine-grained action recognition, robotics, and telepresence.

This is a challenging problem, however. Hands move quickly
as they interact with the world, and handling an object, by definition, creates occlusions of 
the hand and/or object from nearly any given point of view.
Moreover, hand-object interaction video is often collected  from first-person (wearable) cameras (e.g., for Augmented Reality applications), generating
a
large degree of unpredictable camera motion.

Of course, one approach is to detect the poses of the hands and objects
separately~\cite{tekin2018real, yuan2018depth}.  However, this ignores
the fact that hand and handled object poses are highly correlated: the
shape of an object usually constrains the types of grasps (hand poses)
that can be used to handle it.  Detecting the pose of the hand can
give cues as to the pose and identity of an object, while the pose of
an object can constrain the pose of the hand that is holding it.
Solving the two problems jointly
can help overcome challenges such as occlusion.
 Recent work~\cite{tekin2019unified, hasson19_obman} proposed
deep learning-based approaches to jointly model the hand and object
poses. We build on this work, showing how to improve performance
by 
more explicitly modeling the physical and anatomical
constraints on hand-object interaction.

\begin{figure}[t]
  \centering
  \setlength{\tabcolsep}{1pt}
  \begin{tabular}{cc}
  \includegraphics[scale=0.070]{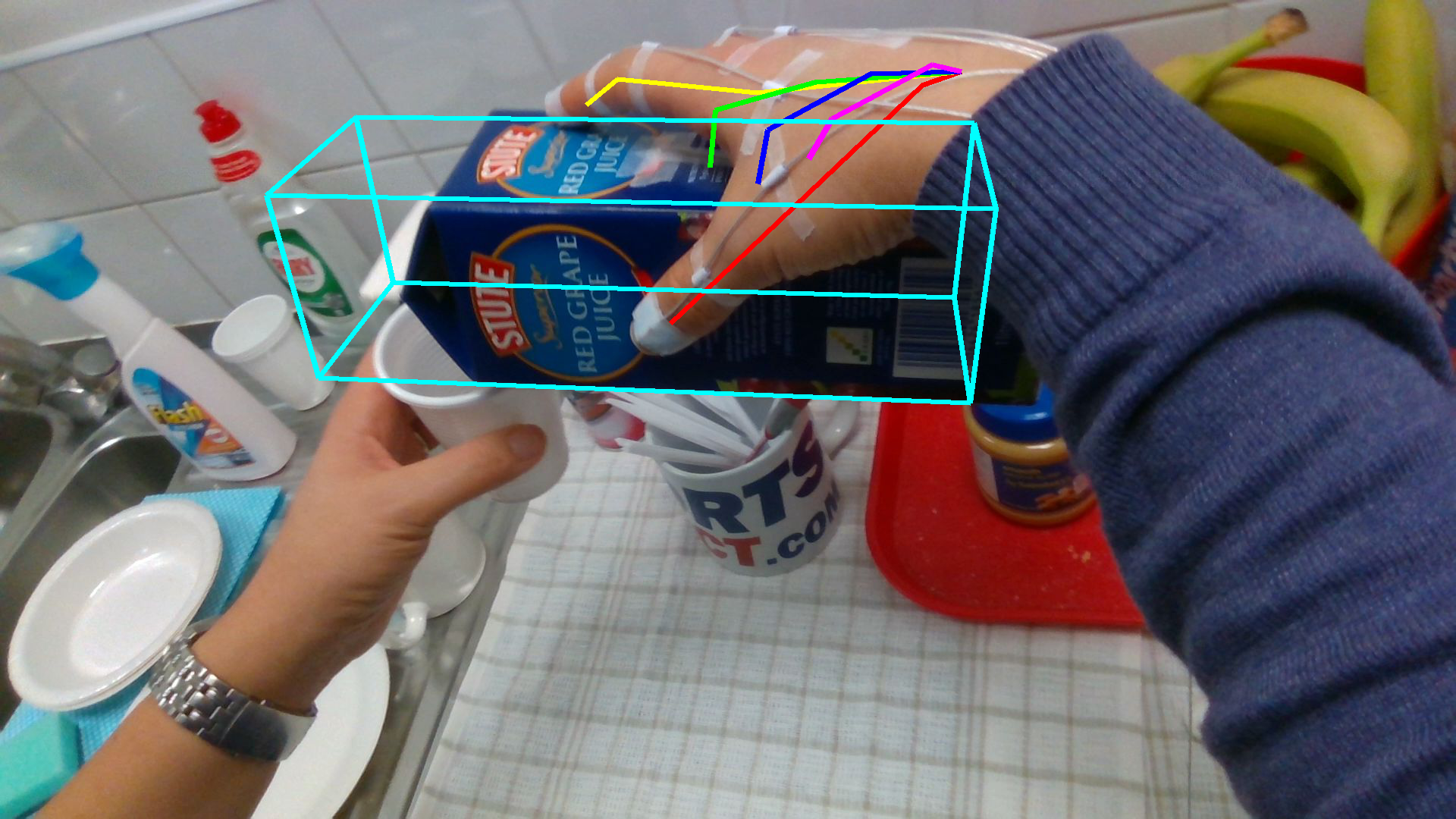} &
  \includegraphics[trim=60 0 0 0,clip,scale=0.20]{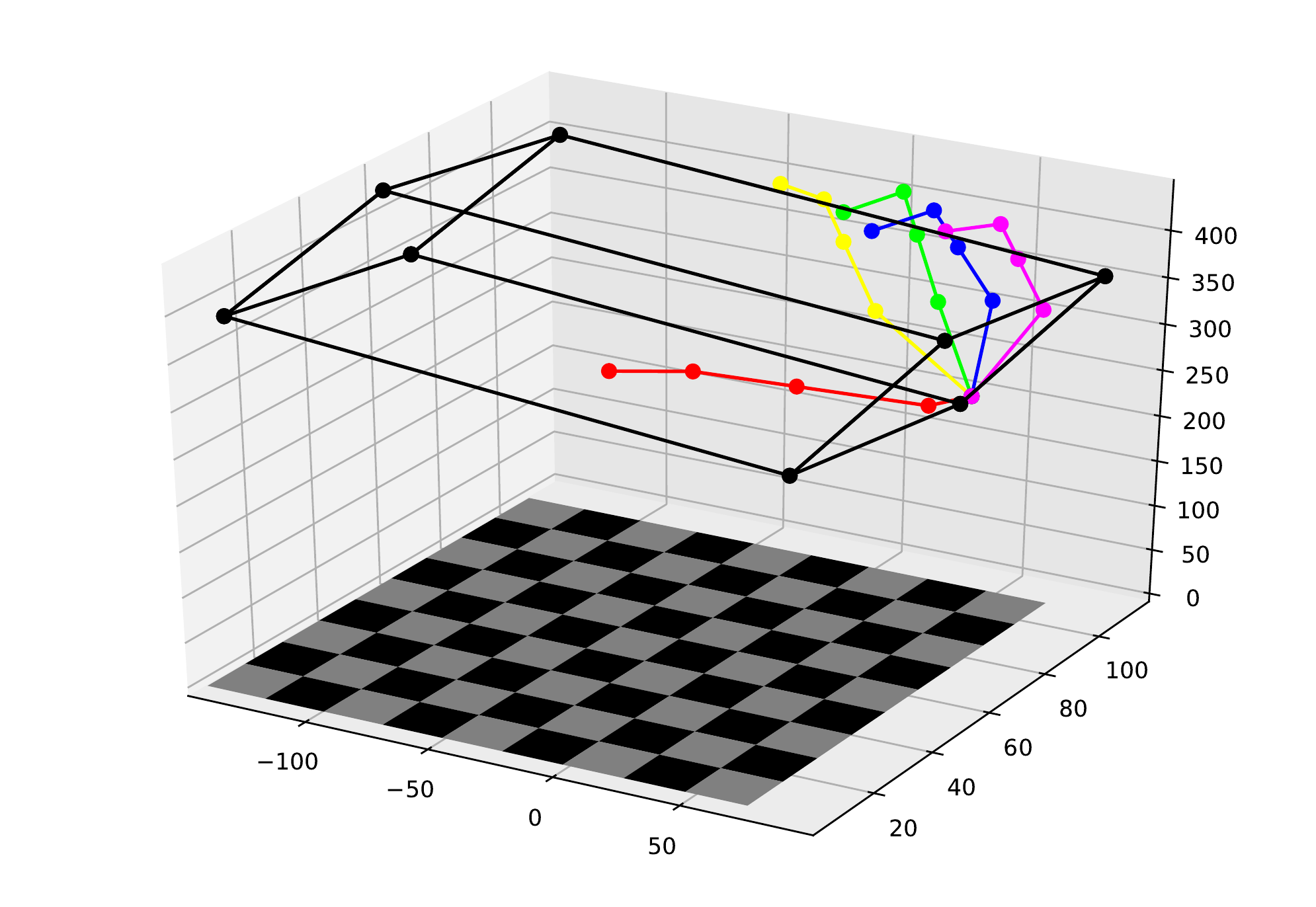}
  \end{tabular}
  \caption{The goal of Hand-Object Pose Estimation (HOPE) is to
    jointly estimate the poses of both the hand and a handled object.
    Our HOPE-Net model can estimate the 2D and
    3D hand and object poses in real-time, given a single image.}
  \label{fig:sample}
\end{figure}

We propose to do this using graph convolutional neural networks.
Given their ability to learn effective representations of graph-structured data, graph convolutional neural networks have recently
received much attention in computer vision. 
Human hand and body pose estimation problems are particularly amenable to
graph-based techniques since they can naturally model the skeletal and kinematic constraints 
between joints and body parts.
Graph convolution can
be used to learn these inter-joint relationships.


In this paper, we show that graph convolution can dramatically increase the
performance of estimating
 3D hand-object pose in real-world
hand-object manipulation videos.
We model hand-object interaction by
representing the hand and object as a single graph. We focus on
estimating 3D hand-object poses from egocentric (first-person) and third-person
monocular color video frames, without requiring any depth
information. Our model first predicts 2D keypoint locations of
hand joints and object boundaries. Then the model jointly
recovers the depth information from the 2D  pose estimates in a
hierarchical manner (Figure~\ref{fig:sample}).

This approach of first estimating in 2D and then ``converting''
to 3D is inspired by the fact that 
detection-based models perform better in detecting 2D hand
keypoints, but in 3D, because of the high degree
of non-linearity and the huge output space, regression-based models
are more popular~\cite{survey}. Our graph convolutional approach allows us
to use a detection-based model to detect the hand keypoints in 2D
(which is easier than predicting 3D coordinates), and then
to accurately convert them to 3D coordinates.
We show that using this graph-based network, we are not
limited to training on only annotated real images, but can instead
pre-train the 2D to 3D network separately with synthetic images rendered from 3D meshes of
hands interacting with objects (\eg ObMan
dataset~\cite{hasson19_obman}). This is very useful for training a
model for hand-object pose estimation as real-world annotated data for
these scenarios is scarce and costly to collect.


In brief, the core contributions of our work are:

\begin{itemize}
\item We propose a novel but lightweight deep learning framework, HOPE-Net, 
which can predict 2D and 3D coordinates of hand
  and hand-manipulated object in real-time. Our model accurately
  predicts the hand and object pose from single RGB images.
\item
We introduce the Adaptive Graph U-Net, a graph convolution-based
neural network to convert 2D hand and object poses to 3D 
 with novel graph convolution, pooling, and unpooling
layers. The new formulations of these layers make it more stable and
robust compared to the existing Graph U-Net~\cite{gao2019graph}
model.

\item
Through extensive experiments, we show that our approach can
outperform the state-of-the-art models for joint hand and object 3D
pose estimation tasks while still running in real-time.

\end{itemize}

\section{Related Work}
Our work is related to two main lines of research: joint hand-object pose prediction
models and graph convolutional networks for understanding
graph-based data.

\paragraph{Hand-Object Pose Estimation.}
Due to the strong relationship between hand pose and the shape of a
manipulated object, several papers have studied joint estimation of
both hand and object pose. Oikonomidis
\etal~\cite{oikonomidis2011full} used hand-object interaction as
context to better estimate the 2D hand pose from multiview
images. Choi \etal~\cite{choi2017robust} trained two networks, one
object-centered and one hand-centered, to capture information from
both the object and hand perspectives, and shared information between
these two networks to learn a better representation for predicting 3D
hand pose. 
Panteleris \etal~\cite{plastira3d} generated 3D hand pose
and 3D models of unknown objects based on hand-object interactions and
depth information. Oberweger \etal~\cite{oberweger2019generalized}
proposed an iterative approach by using Spatial Transformer Networks
(STNs) to separately focus on the manipulated object and the hand to
predict their corresponding poses. Later they estimated the hand and
object depth images and fused them using an inverse STN. The
synthesized depth images were used to refine the hand and object pose
estimates. 
Recently, Hasson \etal~\cite{hasson19_obman} showed that
by incorporating physical constraints, two separate networks
responsible for learning object and hand representations can be
combined to generate better 3D hand and object shapes. Tekin
\etal~\cite{tekin2019unified} proposed a single 3D YOLO model to
jointly predict the 3D hand pose and object pose from a single RGB
image.

\paragraph{Graph Convolution Networks.}
Graph convolution networks allow learning high-level representations
of the relationships between the nodes of graph-based data. Zhao
\etal~\cite{zhao2019semantic} proposed a semantic graph convolution
network for capturing both local and global relationships among human
body joints for 2D and 3D human pose estimation. Cai
\etal~\cite{cai2019exploiting} converted 2D human joints to 3D by
encoding domain knowledge of the human body and hand joints using a
graph convolution network which can learn multi-scale
representations. Yan \etal~\cite{yan2018spatial} used a graph
convolution network for learning a spatial-temporal representation of
human body joints for skeleton-based action recognition. Kolotouros
\etal~\cite{kolotouros2019convolutional} showed that graph
convolutional networks can be used to extract 3D human shape and pose
from a single RGB image, while Ge \etal~\cite{ge20193d} used them to
generate complete 3D meshes of hands from images. Li
\etal~\cite{li2019skeleton} used graph convolutional networks for
skeleton-based action recognition, while  Shi
\etal~\cite{shi2019adaptive,shi2019skeleton} similarly used two stream
adaptive graph convolution.

\begin{figure*}[ht]
  \centering
  \includegraphics[scale=0.5]{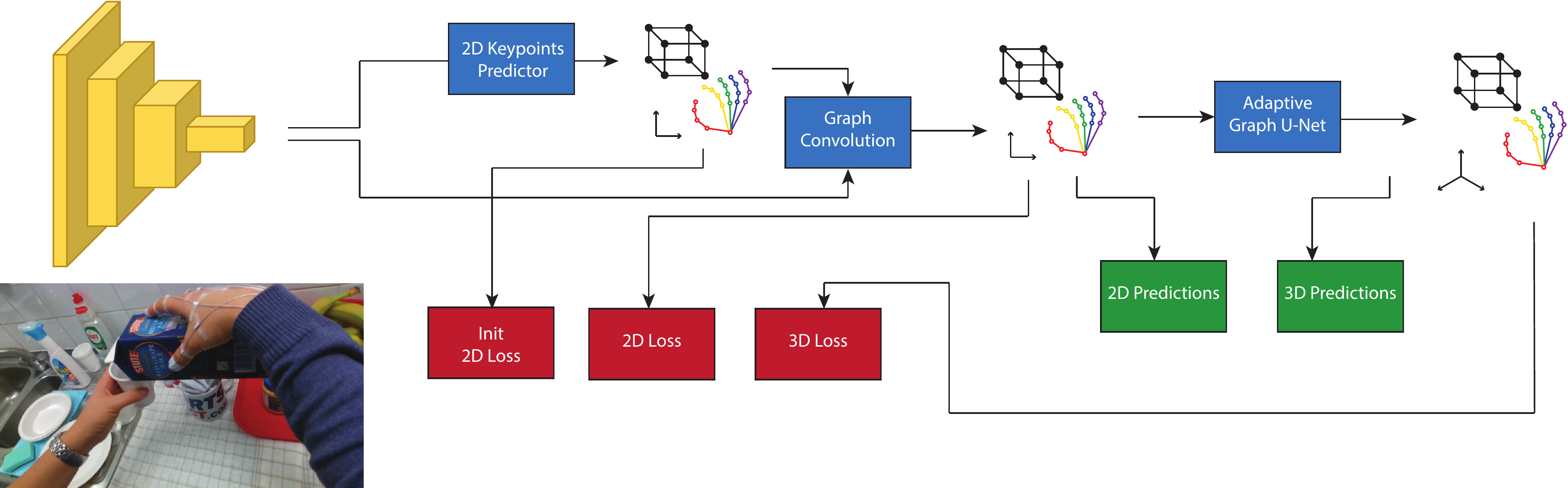}
  \vspace{-5pt}
  \caption{The architecture of HOPE-Net. The model starts
    with ResNet10 as the image encoder and for predicting the
    initial 2D coordinates of the joints and object vertices. The
    coordinates concatenated with the image features used as the
    features of the input graph of a 3 layered graph convolution to
    use the power of neighbors features to estimate the better 2D
    pose. Finally the 2D coordinates predicted in the previous step
    are passed to our Adaptive Graph U-Net to find the 3D coordinates
    of the hand and object.}
  \label{fig:hopenet}
\end{figure*}

Gao \etal~\cite{gao2019graph} introduced the Graph U-Net structure with their proposed
pooling and unpooling layers. But that pooling method did not work well
on graphs with low numbers of edges, such as skeletons or object
meshes. Ranjan \etal~\cite{ranjan2018generating} used fixed pooling
and Hanocka \etal~\cite{hanocka2019meshcnn} used edge pooling to
prevent  holes in the mesh after pooling. In this paper, we
propose a new Graph U-Net architecture with different graph
convolution, pooling, and unpooling. We use an adaptive adjacency
matrix for our graph convolutional layer and   new trainable
pooling and unpooling layers.

\section{Methodology}

We now present HOPE-Net, which consists of
a convolutional neural network for encoding the image and predicting
the initial 2D locations of the hand and object keypoints (hand joints
and tight object bounding box corners), a simple graph convolution
to refine the predicted 2D predictions,
and  a Graph U-Net architecture to convert 2D
keypoints to 3D using a series of graph convolutions, 
poolings, and  unpoolings. 
Figure~\ref{fig:hopenet}
shows an overall schematic of the HOPE-Net architecture.

\subsection{Image Encoder and Graph Convolution}
\label{imageencoder}
For the image encoder, we use a lightweight residual neural
network~\cite{resnet} (ResNet10) 
to help reduce overfitting.
The image
encoder produces a 2048D feature vector for each input
image. Then initial predictions of the 2D coordinates of the
keypoints (hand joints and corners of the object's tight bounding box)
are produced using a fully-connected layer. Inspired by the
architecture of~\cite{Kolotouros_2019_CVPR}, we concatenate these
features with the initial 2D predictions of each keypoint, yielding
a graph with \(2050\) features (\(2048\) image features plus initial estimates of \(x\) and \(y\)) for each node. A
3-layer adaptive graph convolution network is applied to this
graph to use adjacency information and modify the 2D coordinates of
the keypoints.  In the next section, we explain the adaptive graph
convolution  in depth.  The concatenation of the
image features to the predicted \(x\) and \(y\) of each keypoint
forces the graph convolution network to modify the 2D coordinates
conditioned on the image features as well as the initial prediction of
the 2D coordinates. These final 2D coordinates of the hand and object
keypoints are then passed to our adaptive Graph U-Net, a graph
convolution network using adaptive convolution, pooling, and unpooling
to convert 2D coordinates to 3D.

\begin{figure*}[ht]
  \centering
  \includegraphics[scale=1]{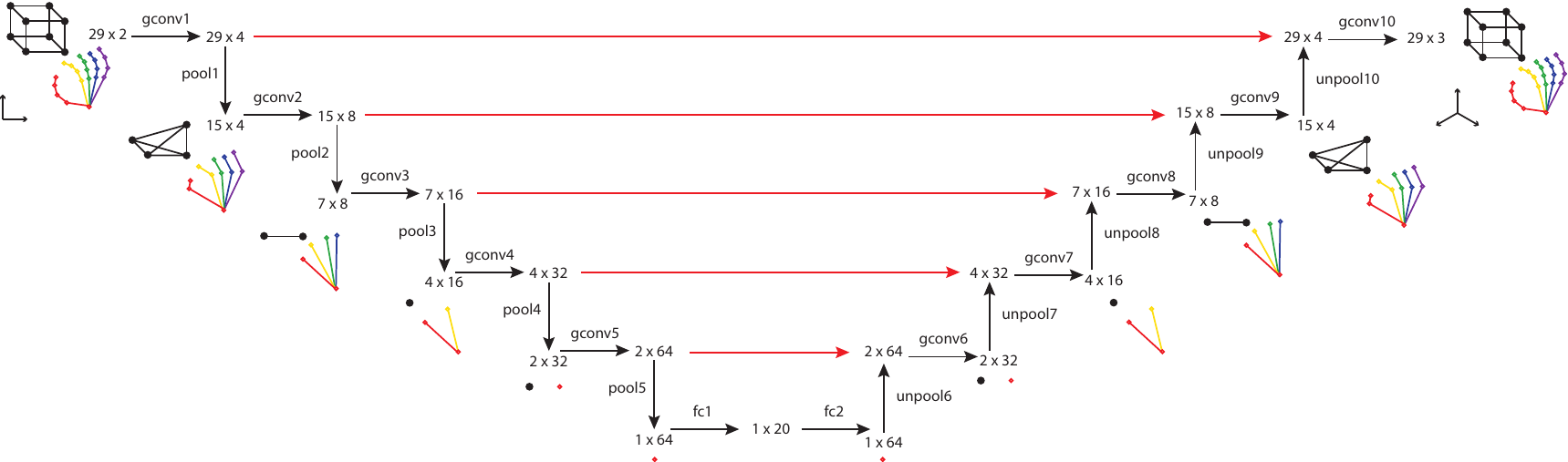}
  \vspace{-5pt}
  \caption{A schematic of our Adaptive Graph U-Net architecture, which is used to
    estimate 3D coordinates from 2D coordinates. In each of the pooling
    layers, we roughly cut the number of nodes in half, while 
    in each unpooling layer, we double the number of nodes in the graph. The
    red arrows in the image are the skip layer features which are
    passed to the decoder to be concatenated with the unpooled
    features.}
  \label{fig:graphunet}
\end{figure*}

\subsection{Adaptive Graph U-Net}
\label{graph}
In this section, we explain our graph-based model which predicts 3D
coordinates of the hand joints and object corners based on 
predicted 2D coordinates. In this network, we simplify the input graph
by applying graph pooling in the encoding part, and in the decoding
part, we add those nodes again with our graph unpooling layers. Also,
similar to the classic U-Net~\cite{ronneberger2015unet},
we use skip connections and concatenate features from the encoding stage to
features of the decoding stage in each decoding graph
convolution. With this architecture we are interested in training a
network which simplifies the graph to obtain global features of the
hand and object, but also tries to preserve local features via
the skip connections from the encoder to the decoder layers. Modeling the
HOPE problem as a graph helps use neighbors to predict
more accurate coordinates and also to discover the relationship between
hands and objects.

The Graph U-Net concept was previously introduced by Gao
\etal~\cite{gao2019graph}, but our network layers, \ie graph
convolution, pooling, and unpooling, are significantly
different. We found that the sigmoid function in the
pooling layer of~\cite{gao2019graph} (gPool) can cause the gradients to
vanish and to not update the picked nodes at all. We thus use a
fully-connected layer to pool the nodes  and updated our
adjacency matrix in the graph convolution layers, using the adjacency
matrix as a kernel we apply to our graph.  Moreover, Gao \etal's
gPool~\cite{gao2019graph} removes the vertices and all the edges
connected to them and does not have a procedure to reconnect the
remaining vertices.
This approach may not be problematic for dense graphs (\eg Citeseer~\cite{kipf2017semi}) in which removing a node and its edges will not change the connectivity of the graph. But in graphs with sparse adjacency matrices, such as when the graph is a mesh or a hand or body skeleton, removing one node and its edges may cut the graph into several isolated subgraphs and destroy the connectivity, which is the most important feature of a graph convolutional neural network. 
Using an adaptive graph convolution neural network, we avoid this problem as the network finds the connectivity of the nodes after each pooling layer.

Below we explain the three components of our network, graph
convolution, pooling, and unpooling layers, in detail. The
architecture of our adaptive Graph U-Net is shown in
Figure~\ref{fig:graphunet}.

\subsubsection{Graph Convolution}

The core part of a graph convolutional  network is the
implementation of the graph convolution operation.  We
implemented our convolution based on the \textit{Renormalization
  Trick} mentioned in~\cite{kipf2017semi}: the output
features of a graph convolution layer for an input graph with \(N\)
nodes, \(k\) input features, and \(\ell\) output features for each node is computed as,
\begin{equation}
Y = \sigma(\tilde{A}XW),
\end{equation}
where \(\sigma\) is the activation function, \(W \in \mathbb{R}^{k \times \ell}\) is the trainable weights
matrix, \(X \in \mathbb{R}^{N \times k}\) is the matrix of input
features, and \(\tilde{A} \in \mathbb{R}^{N \times N}\) is the
row-normalized adjacency matrix of the graph,
\begin{equation}
\tilde{A} = \hat{D}^{-\frac{1}{2}}\hat{A}\hat{D}^{-\frac{1}{2}},
\end{equation}
where \(\hat{A} = A + I\) and \(\hat{D}\) is the diagonal node degree
matrix. \(\tilde{A}\) simply defines the extent to which each node uses other
nodes' features. So \(\tilde{A}X\) is the new
feature matrix in which each node's features are the averaged features
of the node itself and its adjacent nodes. Therefore, to effectively formulate
the HOPE problem in this framework, an effective adjacency
matrix is needed.

Initially, we tried using the adjacency matrix defined by the
kinematic structure of the hand skeleton and the object bounding box
for the first layer of the network.  But we found it was better to
allow the network to learn the best adjacency matrix. Note that this is
no longer strictly an adjacency matrix in the strict sense, but more
like an ``affinity'' matrix where nodes can be connected by weighted
edges to many other nodes in the graph.  An adaptive graph convolution
operation updates the adjacency matrix (\(A\)), as well as the weights
matrix (\(W\)) during the backpropagation step. This approach allows
us to model subtle relationships between joints which are not
connected in the hand skeleton model (\eg strong
relationships between finger tips despite not being
physically connected).

We use ReLU as the activation function for the graph convolution
layers. Also we found that the network trains faster and generalizes
better if we do not use either Batch~\cite{ioffe2015batch} or Group
Normalization~\cite{wu2018group}.

\subsubsection{Graph Pooling}
As mentioned earlier, we did not find gPool~\cite{gao2019graph}
helpful in our problem: the sigmoid function's weaknesses
are well-known~\cite{krizhevsky2012imagenet, nair2010rectified}
and the use of sigmoid in the pooling step created very small
gradients during backpropagation. This caused the network not to
update the randomly-initialized selected pooled nodes throughout the entire
training phase, and lost the advantage of the trainable pooling layer.

To solve this problem, we use a fully-connected layer and apply it on
the transpose of the feature matrix. This fully-connected works as a
kernel along each of the features and outputs the desired number of
nodes. Compared to gPool, we found this module updated
very well during training. Also due to using an adaptive graph
convolution, this pooling does not fragment the graph into pieces.


\subsubsection{Graph Unpooling}

The unpooling layer used in our Graph U-Net is also different from
Gao \etal's gUnpool~\cite{gao2019graph}. That approach
adds the pooled nodes to the graph with empty features and uses the
subsequent graph convolution to fill those features. Instead, we use a
transpose convolution approach in our unpooling layer. Similar to our
pooling layer, we use a fully-connected layer and applied it on the
transpose matrix of the features to obtain the desired number of output
nodes, and then transpose the matrix again.


\subsection{Loss Function and Training the Model}
Our loss function for training the model has three parts. We first
calculate the loss for the initial 2D coordinates predicted by 
ResNet (\(\mathcal{L}_{init 2D}\)). We then add this loss to that
calculated from the predicted 2D and 3D coordinates
(\(\mathcal{L}_{2D}\) and \(\mathcal{L}_{3D}\)),
\begin{equation}
\mathcal{L} = \alpha \mathcal{L}_{init 2D} + \beta \mathcal{L}_{2D} + \mathcal{L}_{3D},
\end{equation}
where we set \(\alpha\) and \(\beta\) to \(0.1\) to bring the 2D error
(in pixels) and 3D error (in millimeters) into a similar range. For
each of the loss functions, we used Mean Squared Error.
%
%

\section{Results}

We now describe our experiments and
report results for hand-object pose estimation.  

\subsection{Datasets}
To evaluate the generality of our hand-object pose estimation method,
we used two datasets with very different contexts:
First-Person Hand Action
Dataset~\cite{FirstPersonActionDataset}, which has videos captured from
egocentric (wearable) cameras, 
and
HO-3D~\cite{hampali2019honnotate}, which was captured from third-person views.
We also used a third dataset of synthetic images, ObMan~\cite{hasson19_obman}, for
pre-training.

\label{sec:dataset}

\textbf{First-Person Hand Action
  Dataset}~\cite{FirstPersonActionDataset} contains first-person
videos of hand actions performed on a variety of objects.  The objects
are \textit{milk}, \textit{juice bottle}, \textit{liquid soap},
and \textit{salt}, and actions include \textit{open}, \textit{close},
\textit{pour}, and \textit{put}. Three-dimensional meshes for the
objects are provided.  Although this is a large dataset, a relatively
small subset of frames (\(21,501\)) include 6D object pose annotations,
with \(11,019\) for training  and \(10,482\) for evaluation.
The annotation provided for each frame is a 6D vector giving 3D
translation and rotation for each of the objects. To fit this
annotation to our graph model, for each object in each frame, we
translate and rotate the 3D object mesh to the pose given by the
annotation, and then compute a tight oriented bounding box (simply PCA
on vertex coordinates). We use the eight 3D
 coordinates of the object box corners as nodes in our graph.


The \textbf{HO-3D Dataset}~\cite{hampali2019honnotate} also
contains hands and handled objects but is quite different because it is
captured from a third-person point-of-view. Hands and objects in these
videos are smaller because they are further from the
camera, and their position is less constrained than in first-person videos 
(where people tend to center their field of view around attended objects).
HO-3D contains \(77,558\) frames annotated with hands and objects, and
was collected with 10 subjects and 10 objects. \(66,034\) frames are
designated as the training set and \(11,524\) are for evaluation.
Hands in the evaluation set of HO-3D are just annotated with the wrist
coordinates and the full hand is not annotated.

\textbf{ObMan}~\cite{hasson19_obman} is a large
dataset of synthetically-generated images of hand-object interactions.
Images in this dataset were produced 
by rendering meshes of hands with selected objects from
ShapeNet~\cite{chang2015shapenet}, using an optimization on the
grasp of the objects.
ObMan contains \(141,550\) training,
\(6,463\) validation, and \(6,285\) 
evaluation frames. 
Despite the  large-scale of the annotated data, we found that models trained with these synthetic images do 
not generalize
well to real images.  Nevertheless, we found it helpful to pretrain our
model on the large-scale data of ObMan, and then fine-tune using real images.

All of these datasets use 21 joints model for hands which contains one joints
for the wrist and 4 joints for each of the fingers.

\subsection{Implementation Details}
\label{sec:implement}

\begin{figure}[t]
  \centering
  \includegraphics[scale=0.5,trim=0 0 0 20,clip]{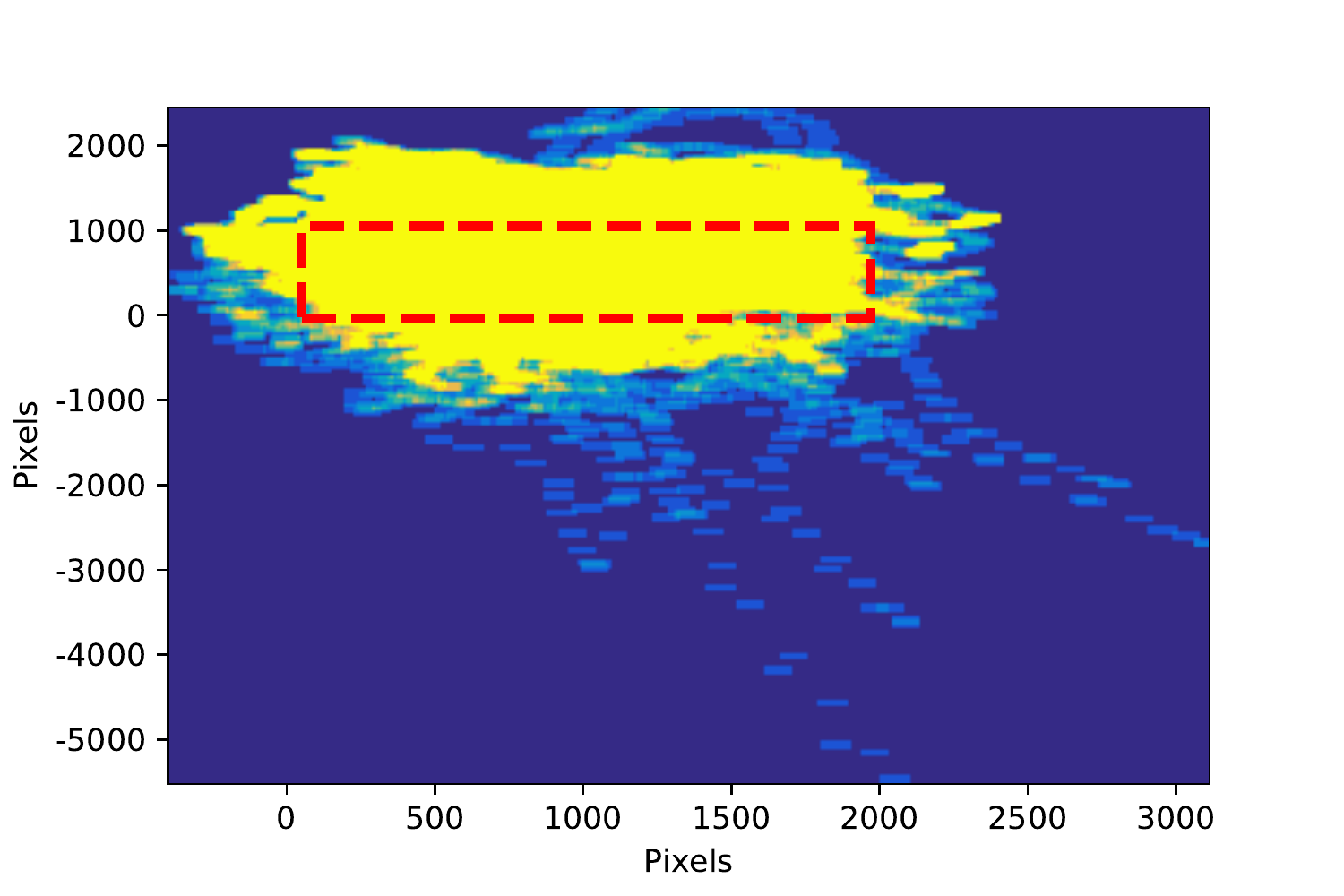}
  \vspace{1pt}
  \caption{Scatter plot of keypoint coordinates in the First Person Hand Action
    dataset. The red dashed rectangle denotes the image frame. Since
    many  points are outside the image boundary, the detection-based
    models did not work well on this dataset.}
  \label{fig:data}
\end{figure}

Because of the nature of first-person video, hands often leave the
field of view, and thus roughly half of the frames in the First-Person
Hand Action dataset have at least one keypoint outside of the frame
(Figure~\ref{fig:data}).  Because of this, we found that
detection-based models are not very helpful in this dataset. Thus we
use a regression-based model to find the initial 2D coordinates. To
avoid overfitting, we use a lightweight ResNet which gave better
generalization. This lightweight model is also  fast, allowing us
to run our model in near real-time. For both datasets, we use the
official training and evaluation splits, and pretrain on
ObMan~\cite{hasson19_obman}.

Since HOPE-Net has different numbers of parameters and complexity, we
train the image encoder and graph parts separately. The 2D to 3D
converter network can be trained separately because
it is not dependent to the annotated image. In addition to the samples
in the FPHA dataset, we augment the 2D points with Gaussian noise ($\mu=0, \sigma=10$)
to help improve
robustness to errors.

For both FPHA and HO-3D datasets
we train the ResNet model with an initial learning rate of \(0.001\)
and multiply it by \(0.9\) every \(100\) steps. We train ResNet for
\(5000\) epochs and the graph convolutional 
network for \(10,000\) epochs, starting from a learning rate of
\(0.001\) and multiplying by \(0.1\) every \(4000\) steps. Finally we
train the model end-to-end for another \(5000\) epochs. All the images
are resized to \(224 \times 224\) pixels and passed to the ResNet. All
learning and inference was implemented in PyTorch.


\subsection{Metrics}
\label{sec:metrics}

Similar to~\cite{tekin2019unified}, we evaluated our model using
percentage of correct pose (PCP) for both 2D and 3D coordinates. In
this metric, a pose is considered  correct if the average
distance to the ground truth pose is less than a 
threshold.

\subsection{Hand-Object Pose Estimation Results}
\label{sec:hope}
\begin{figure}[t]
  \centering
  \includegraphics[scale=0.5,trim=0 0 0 30,clip]{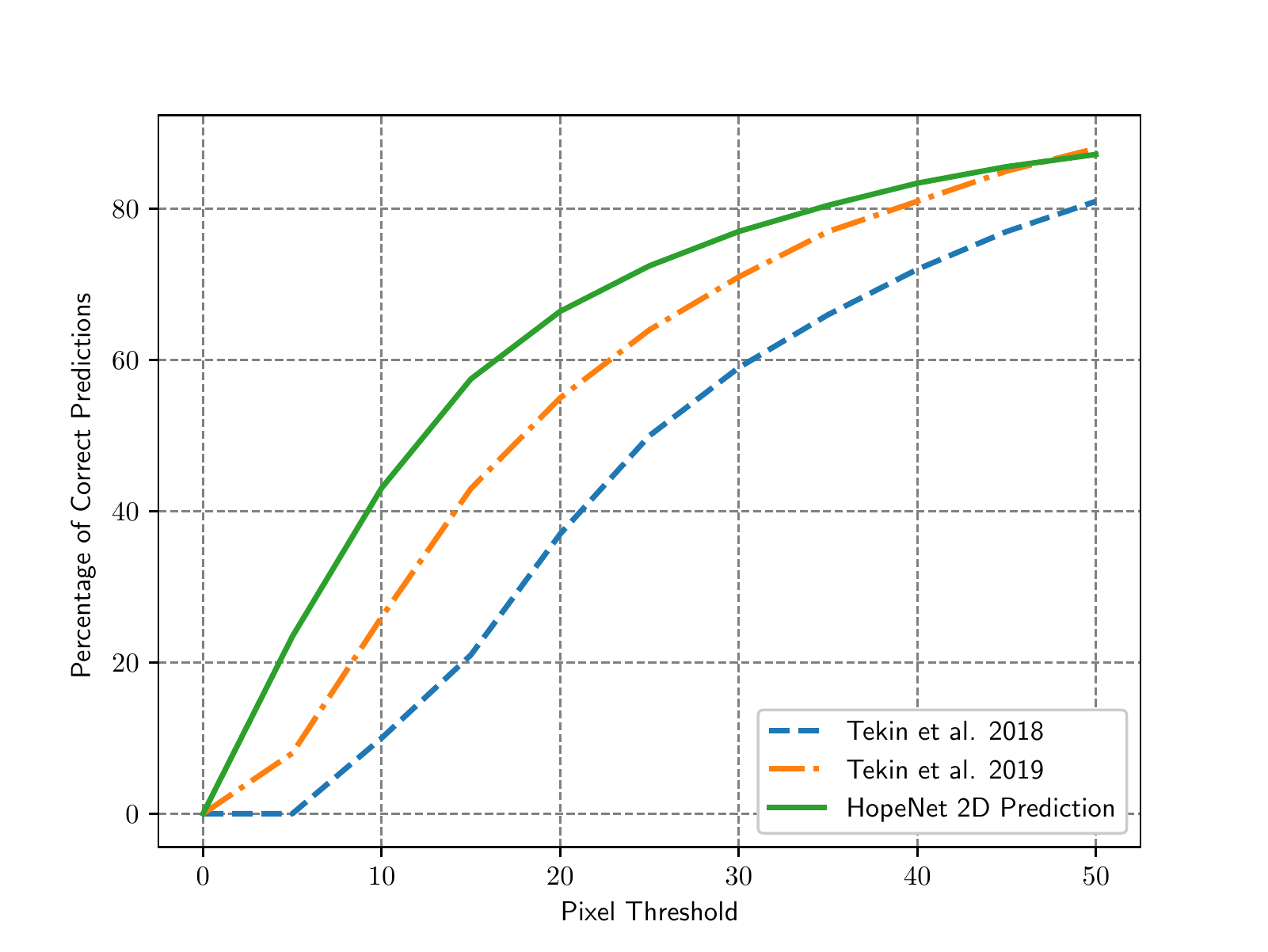}
  \vspace{1pt}
  \caption{The percentage of correct 2D object pose of our model
    on the First-Person Hand Action dataset compared to~\cite{tekin2019unified}
    and~\cite{tekin2018real}. The graph convolutional layers
    helped the model to predict more accurate coordinates.}
  \label{fig:2dgraph}
\end{figure}

We now  report the performance of our model in hand and object
pose estimation on our two  datasets.
%
%
Figure~\ref{fig:2dgraph} presents
the percentage of correct object pose for each pixel threshold on the
First-Person Hand Action dataset.
 As we can see in
this graph, the 2D object pose estimates produced by the HOPE-Net model
outperform the state-of-the-art model of Tekin \etal~\cite{tekin2019unified}
for 2D object pose estimation, even though we do not
use an object locator  and we operate on single frames  without
using temporal constraints.
Moreover,  our
architecture is lightweight and  faster to run.

\begin{figure}[t]
  \centering
  \includegraphics[scale=0.5,trim=0 0 0 30,clip]{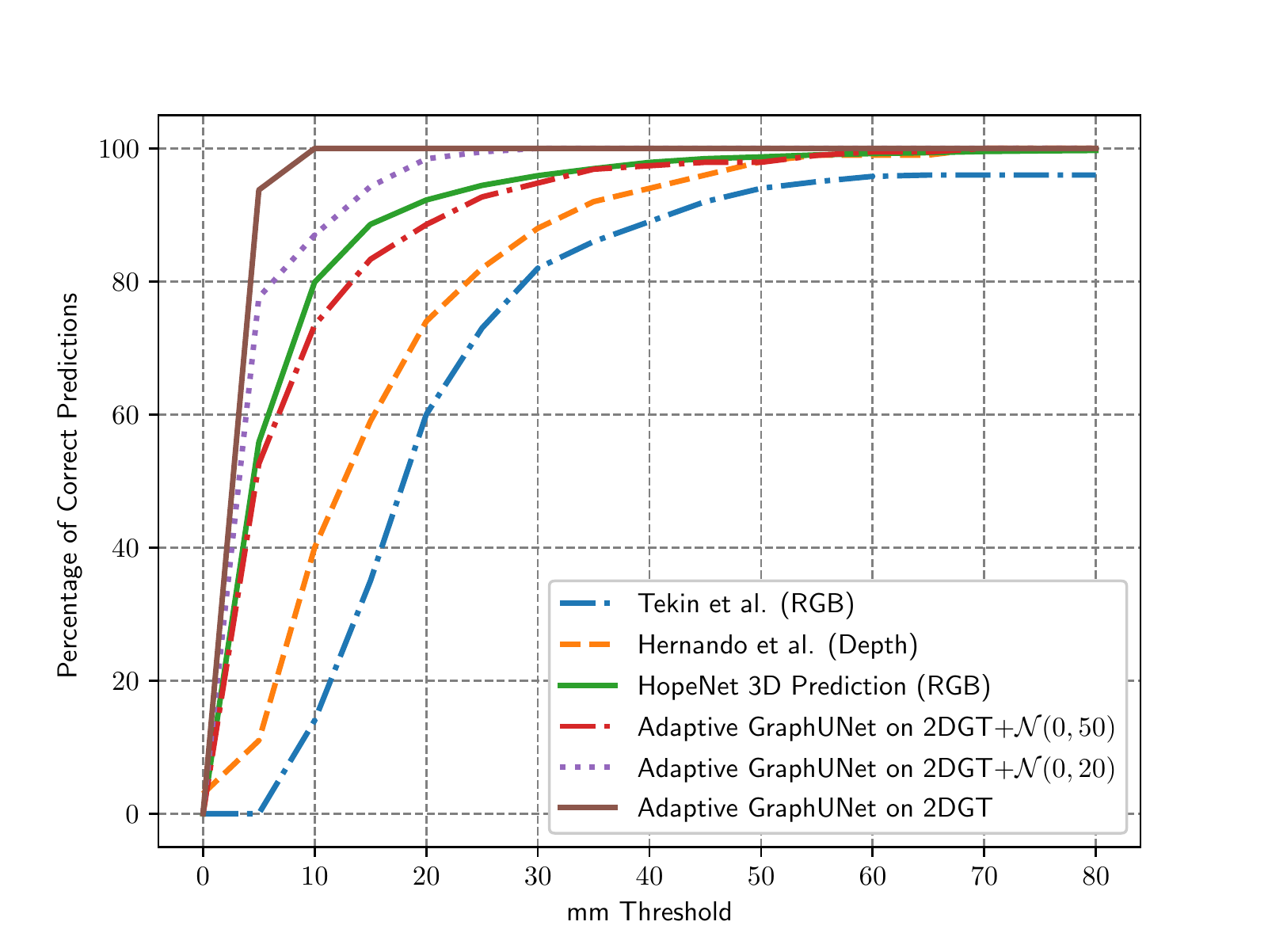}
  \vspace{1pt}
  \caption{The percentage of correct 3D hand pose of our model on
    the First-Person Hand Action dataset compared to the RGB-based technique of~\cite{tekin2019unified} 
    and the depth-based technique of~\cite{FirstPersonActionDataset}. 
    Our model works well on roughly accurate 2D estimates.}
  \label{fig:3dgraph}
\end{figure}

Figure~\ref{fig:3dgraph} presents
 the percentage of correct 3D poses for various thresholds (measured in millimeters) on
the First-Person Hand Action dataset. The results show that
the HOPE-Net model outperforms Tekin \etal's RGB-based
model~\cite{tekin2019unified} and Herando
\etal's~\cite{FirstPersonActionDataset} depth-based model in 3D pose
estimation, even without using an object localizer or temporal
information. 

We also tested our graph model with various other inputs,
including ground truth 2D
coordinates, as well as ground truth 2D coordinates with Gaussian  noise added (with zero mean and
 $\sigma=20$ and $\sigma=50$).
Figure~\ref{fig:3dgraph} presents the results. We note that the graph model is able to effectively
remove the Gaussian noise from the keypoint coordinates. 

\begin{figure*}[ht]
  \centering
  \setlength{\tabcolsep}{1pt}
  \begin{tabular}{cccccc}
  \includegraphics[scale=0.225]{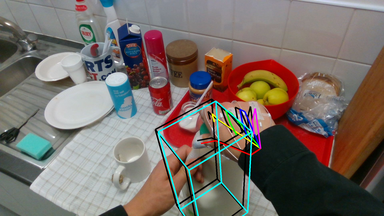} &
  \includegraphics[scale=0.21]{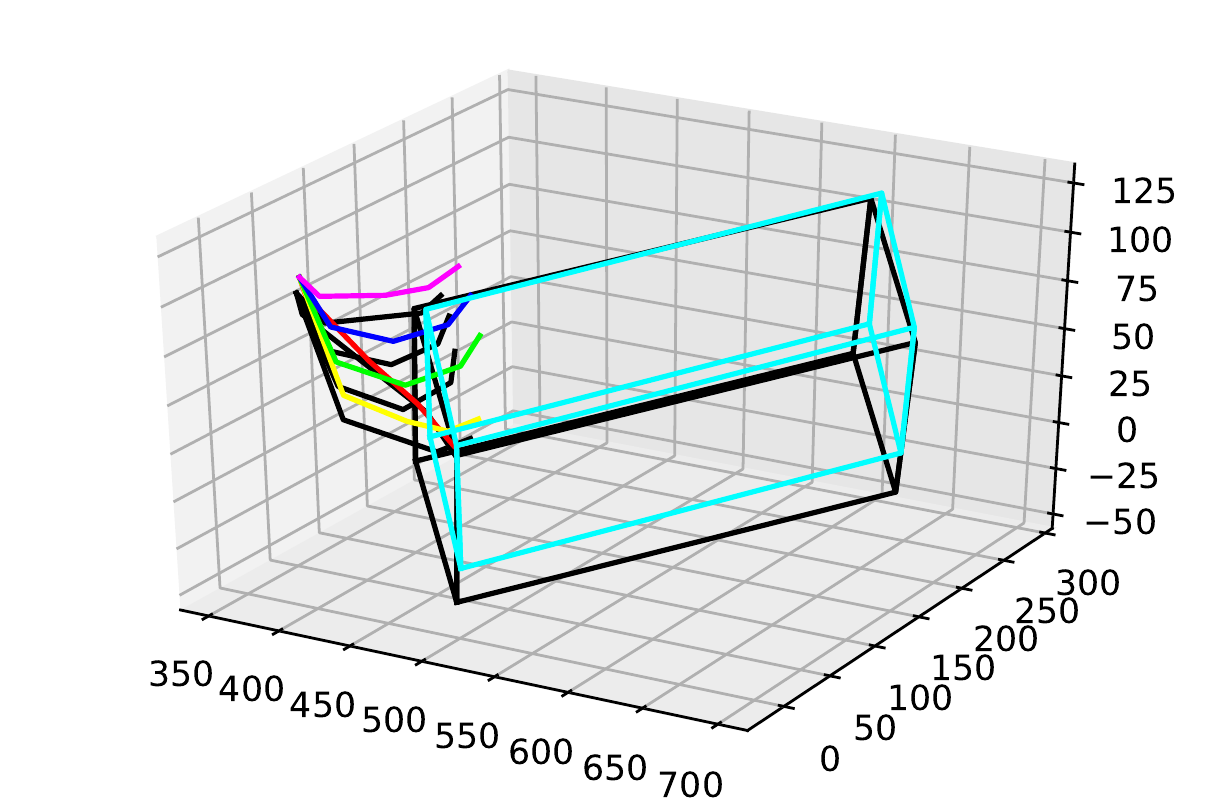} &
  \includegraphics[scale=0.225]{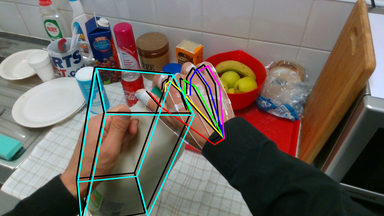} &
  \includegraphics[scale=0.21]{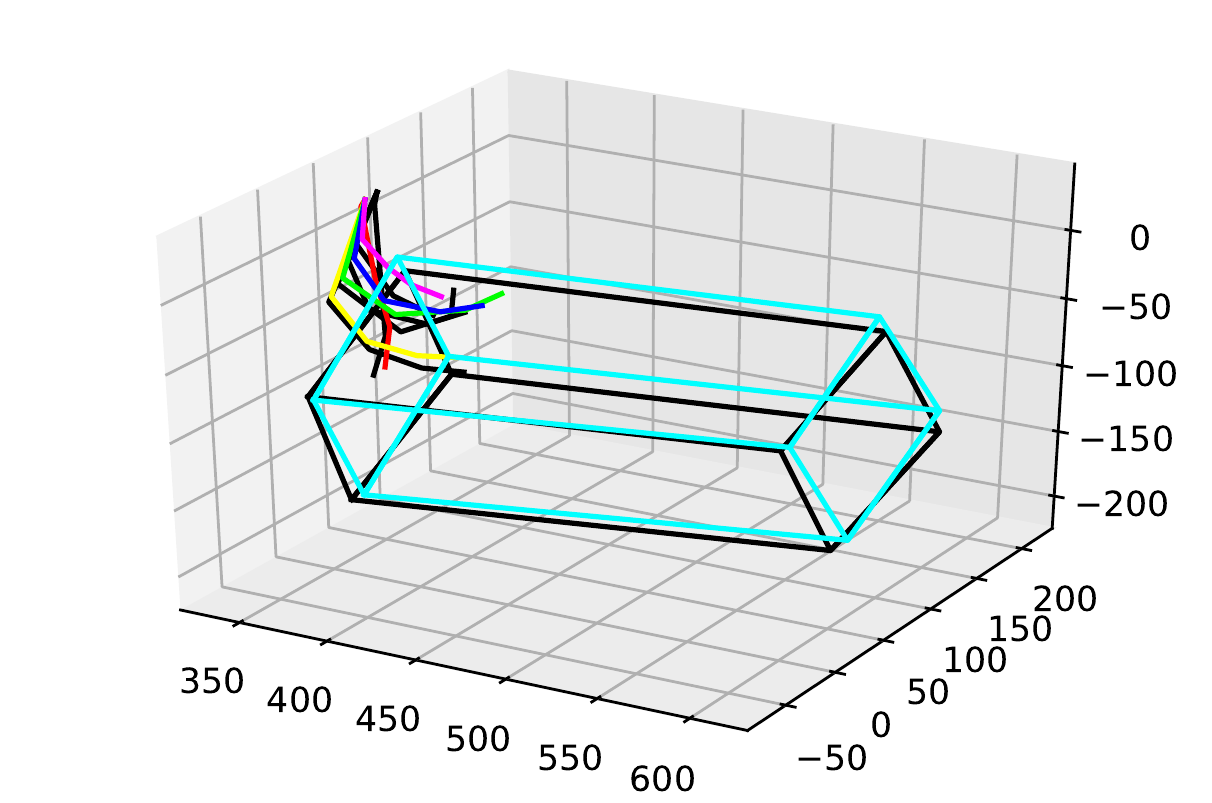} &
  \includegraphics[scale=0.225]{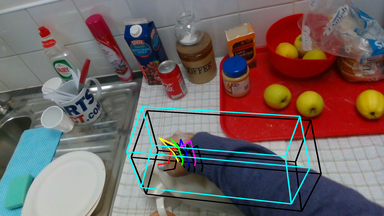} &
  \includegraphics[scale=0.21]{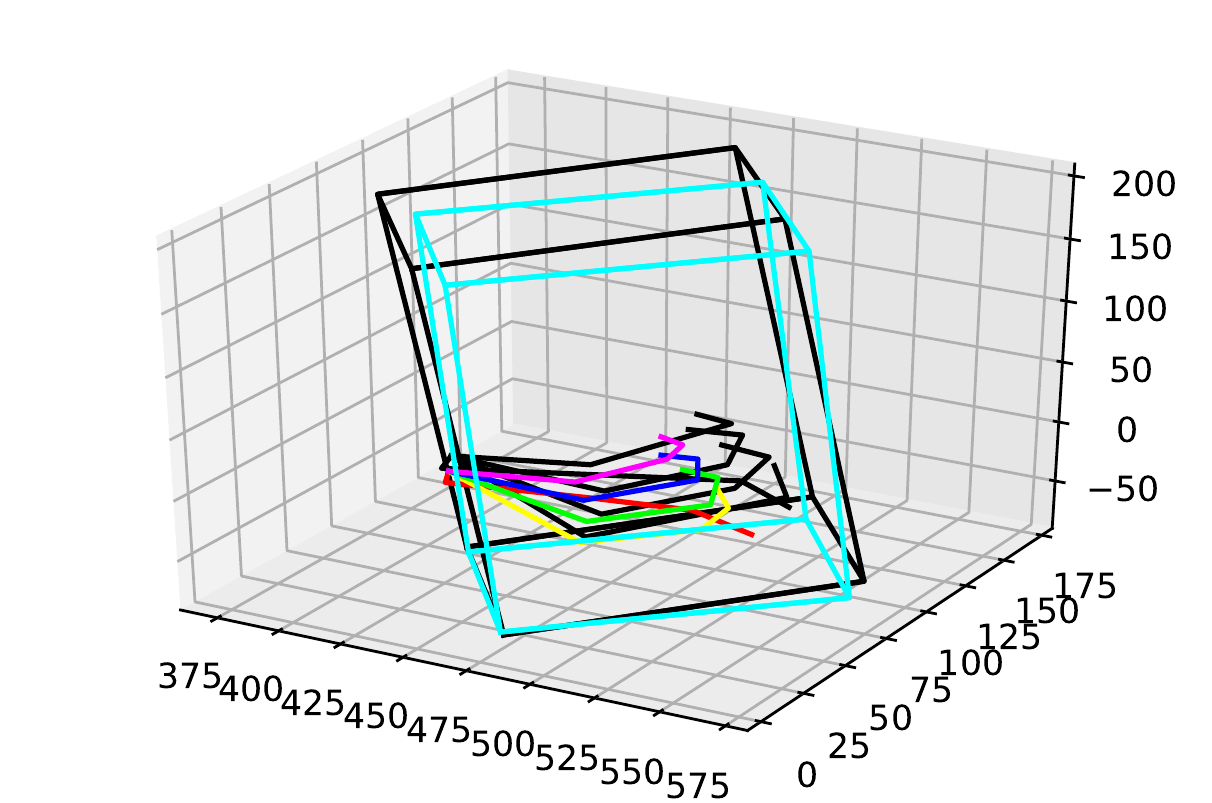} \\
  \includegraphics[scale=0.225]{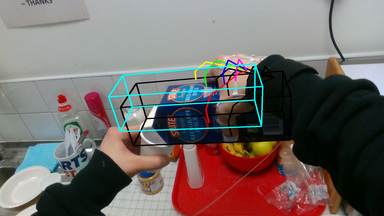} &
  \includegraphics[scale=0.21]{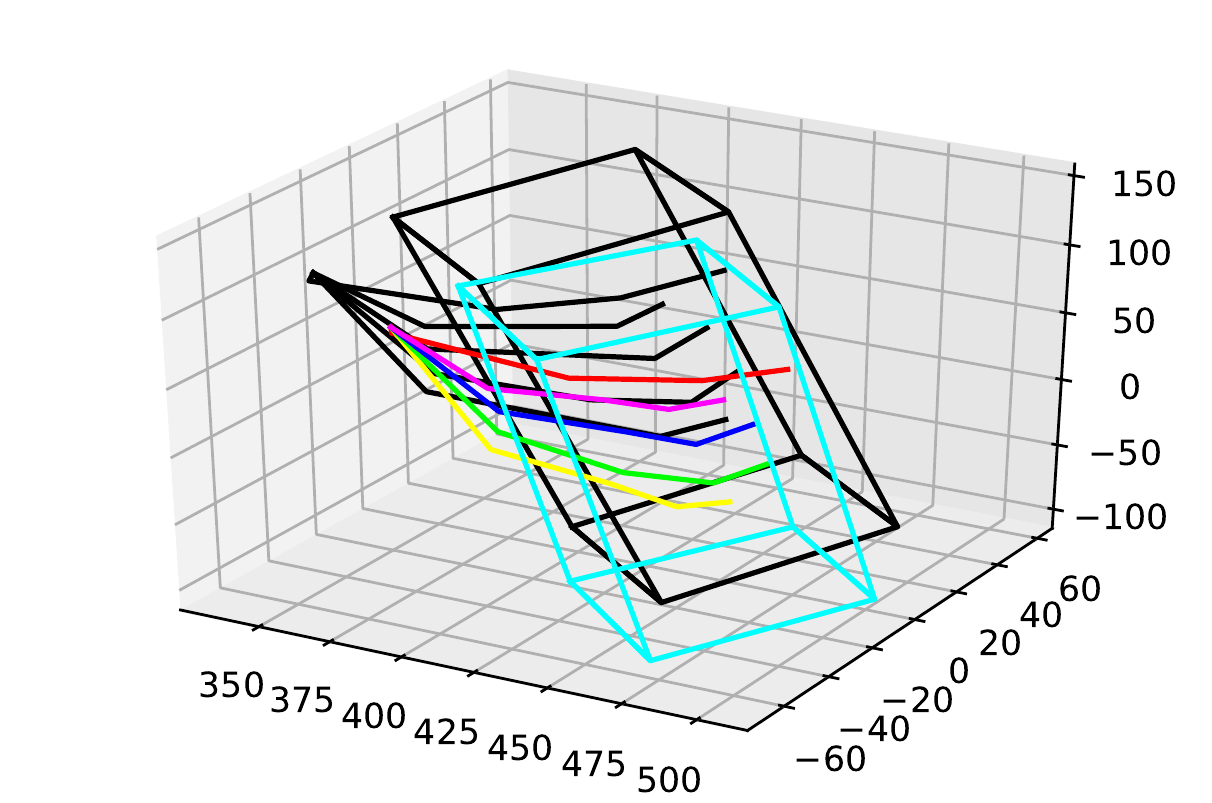} &
  \includegraphics[scale=0.225]{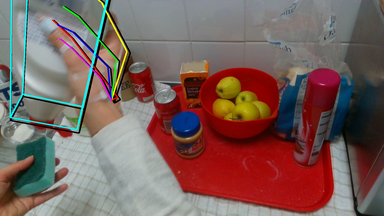} &
  \includegraphics[scale=0.21]{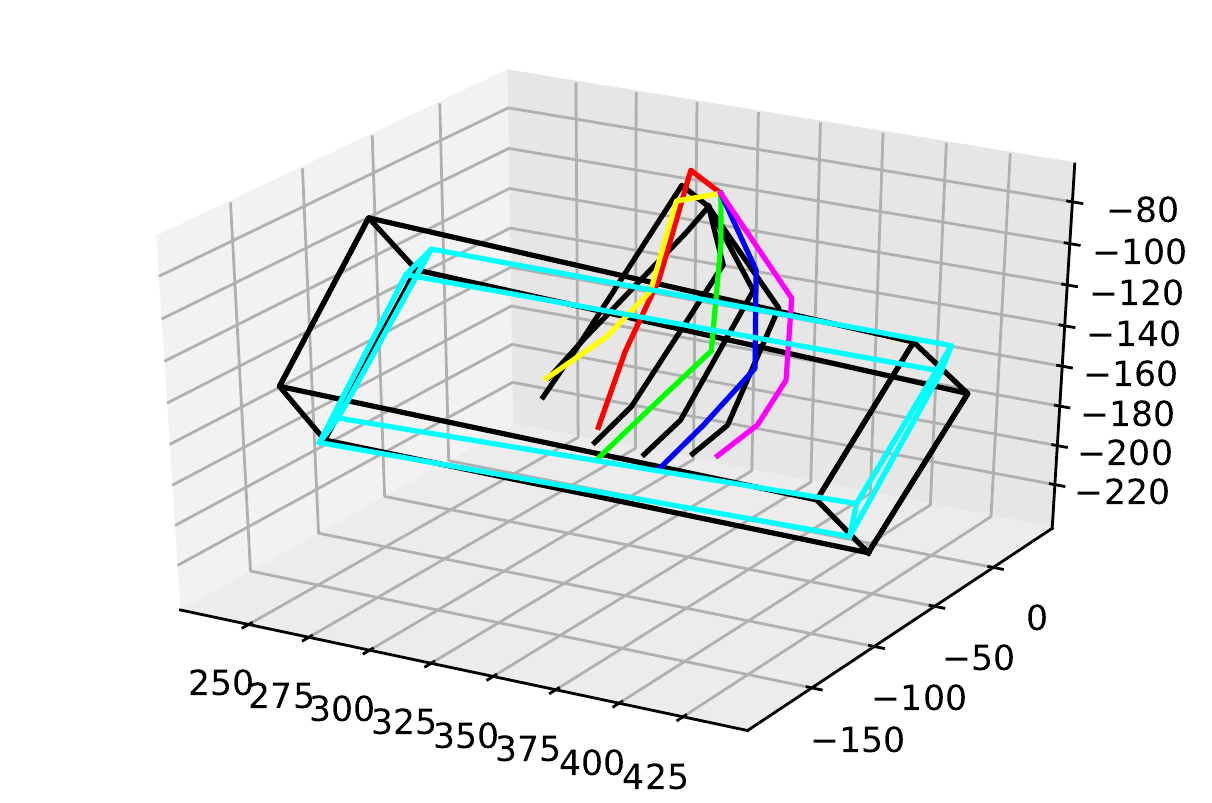} &
  \includegraphics[scale=0.225]{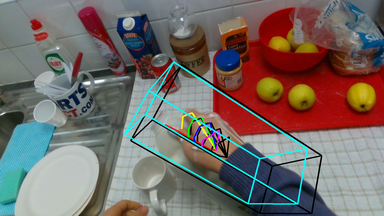} &
  \includegraphics[scale=0.21]{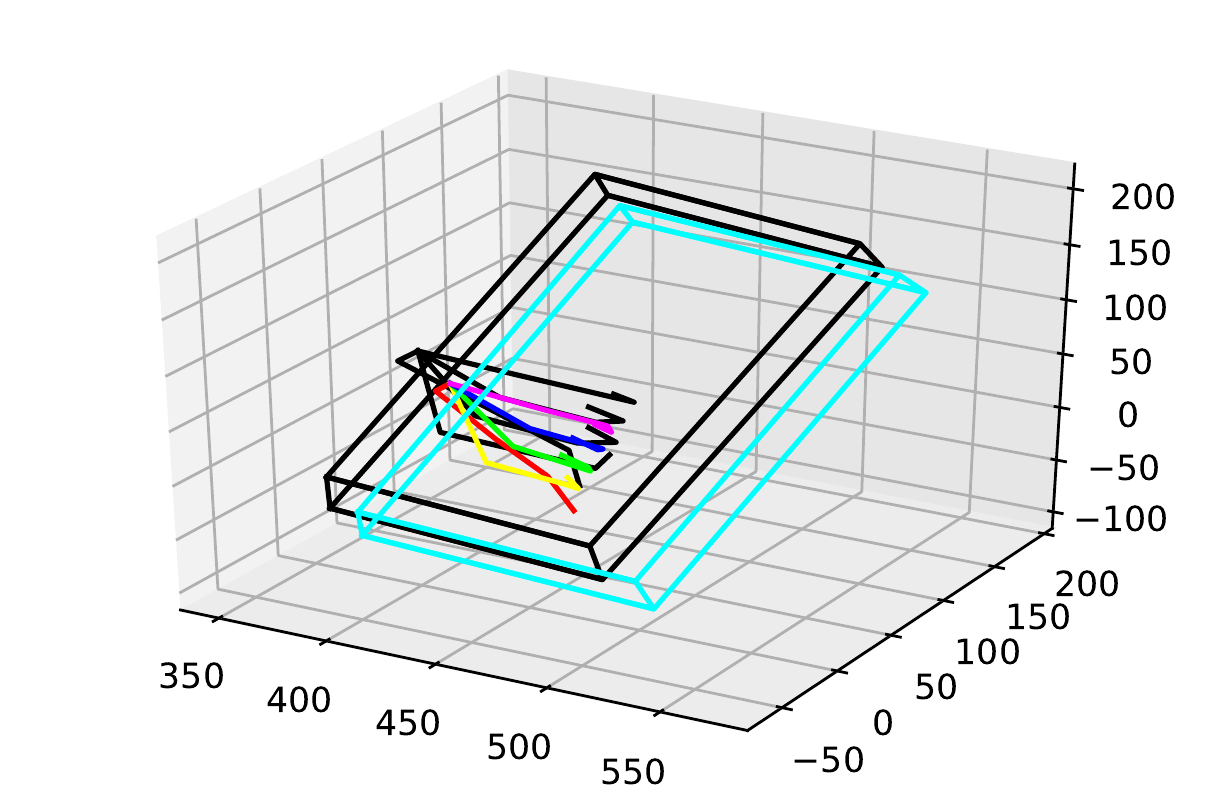} \\
  \includegraphics[scale=0.225]{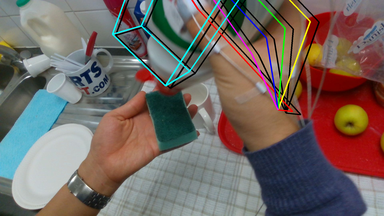} &
  \includegraphics[scale=0.21]{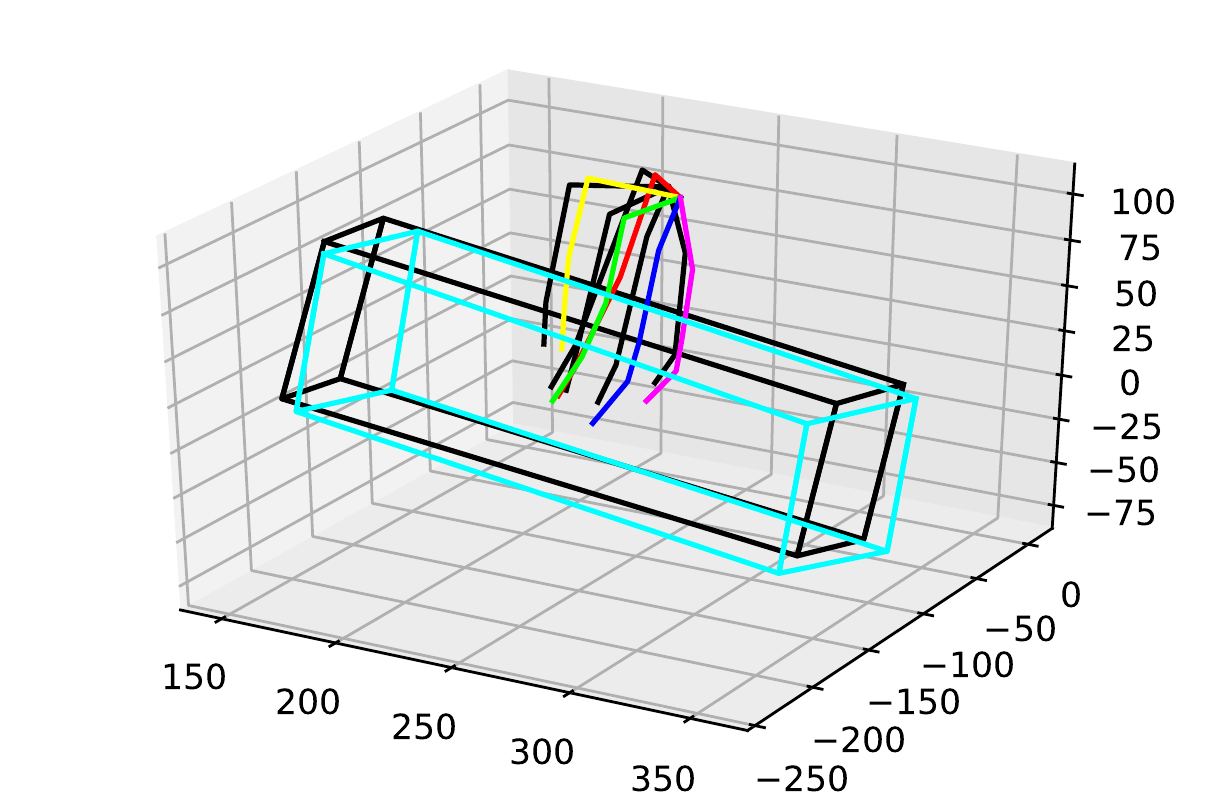} &
  \includegraphics[scale=0.225]{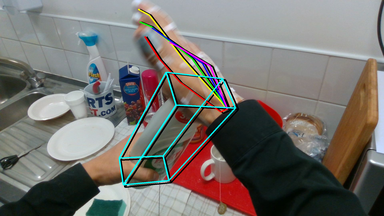} &
  \includegraphics[scale=0.21]{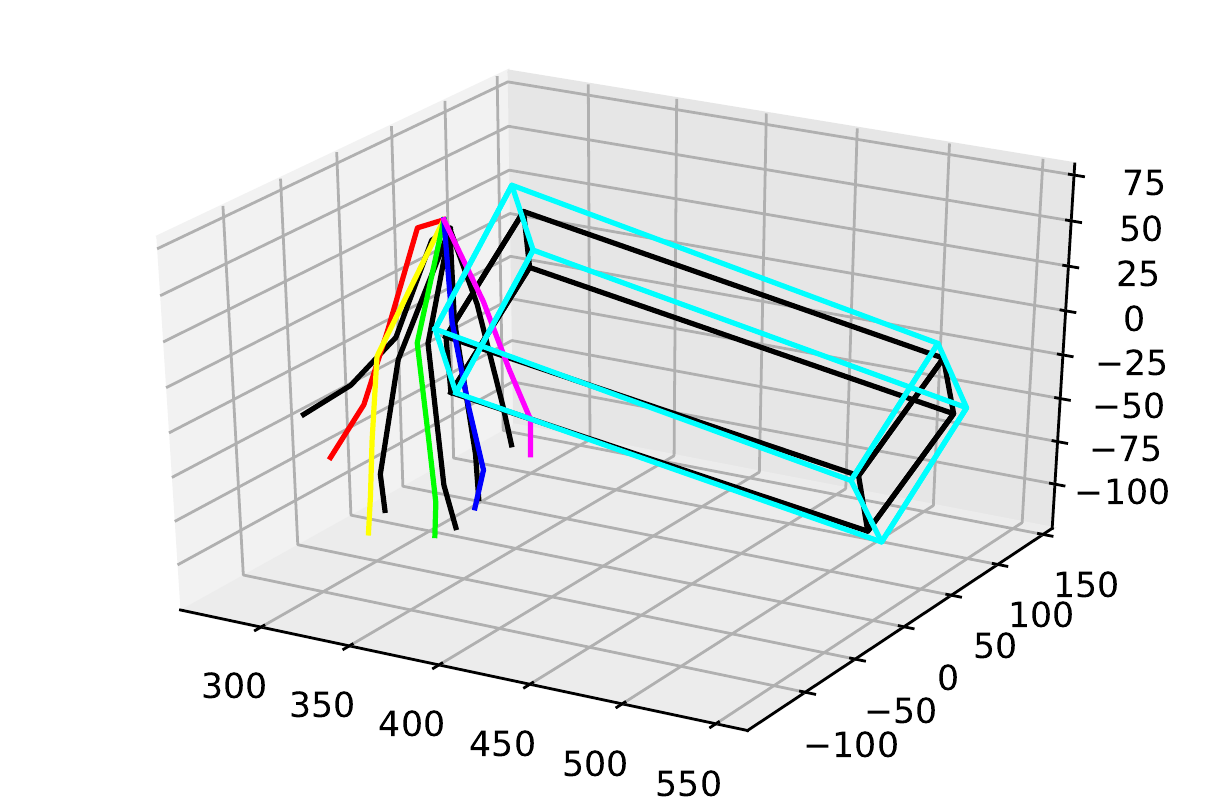} &
  \includegraphics[scale=0.225]{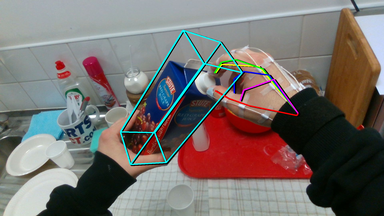} &
  \includegraphics[scale=0.21]{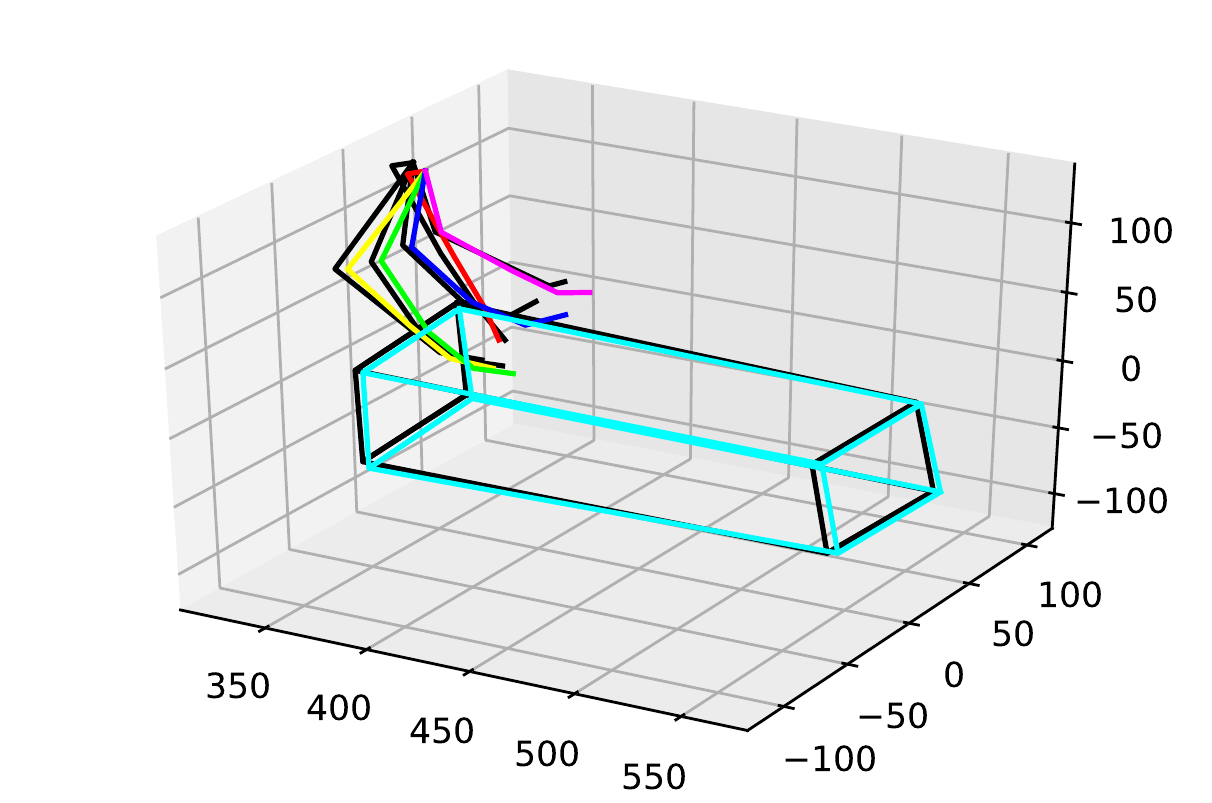} \\
  \includegraphics[scale=0.225]{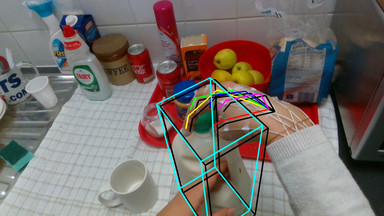} &
  \includegraphics[scale=0.21]{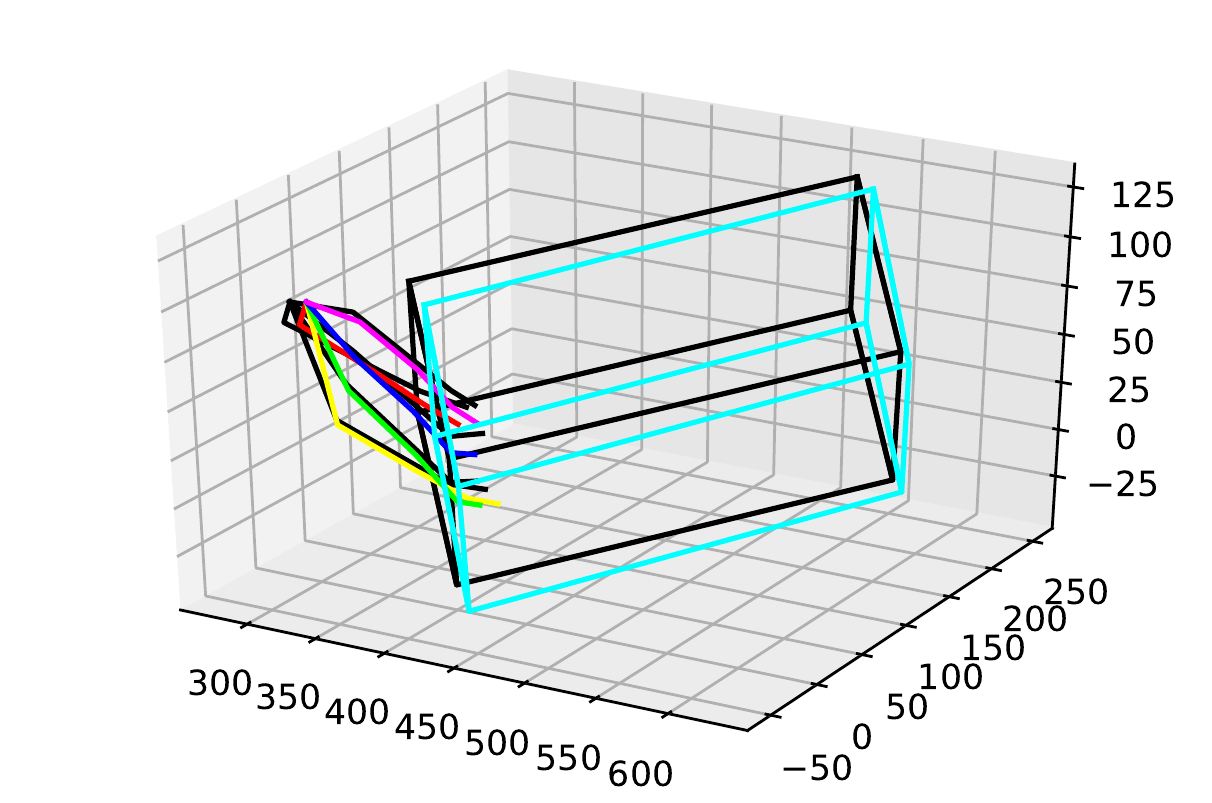} &
  \includegraphics[scale=0.225]{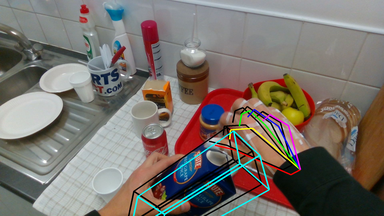} &
  \includegraphics[scale=0.21]{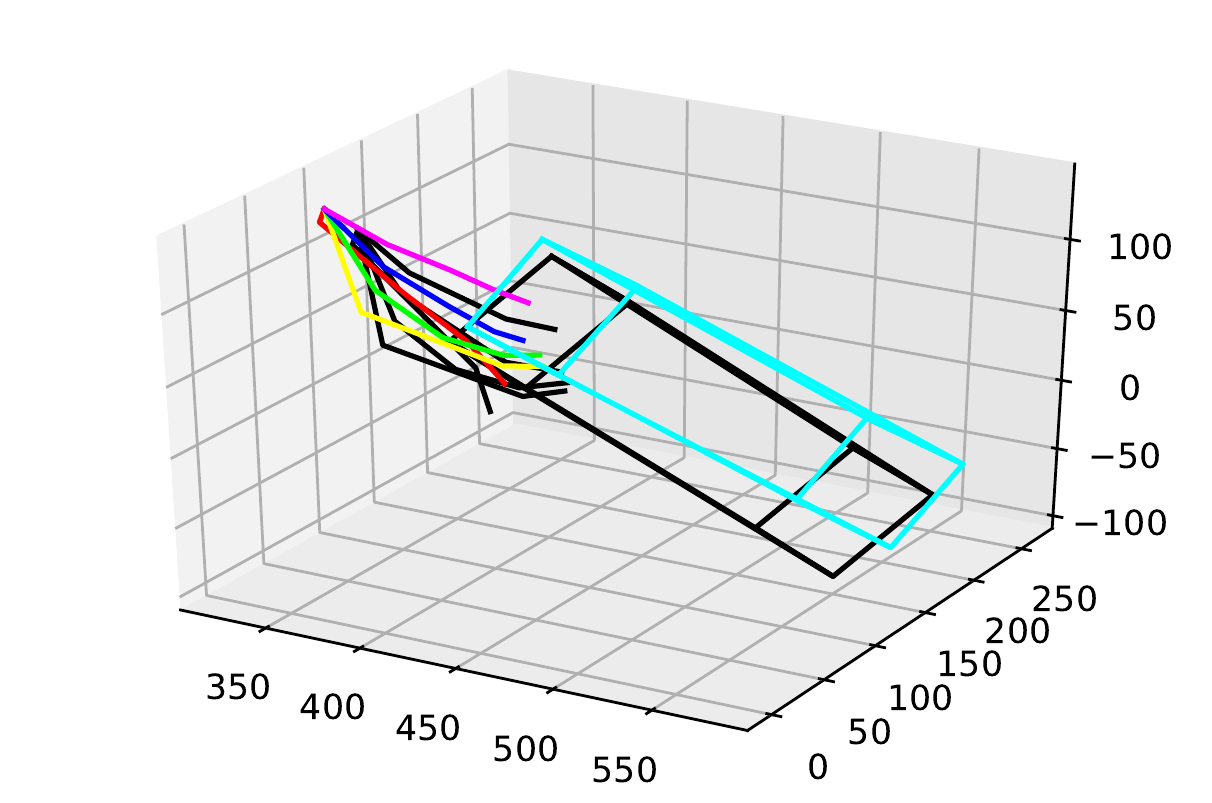} &
  \includegraphics[scale=0.225]{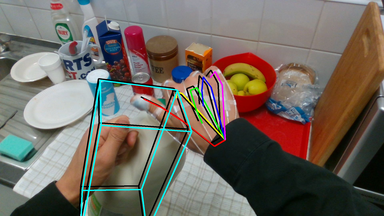} &
  \includegraphics[scale=0.21]{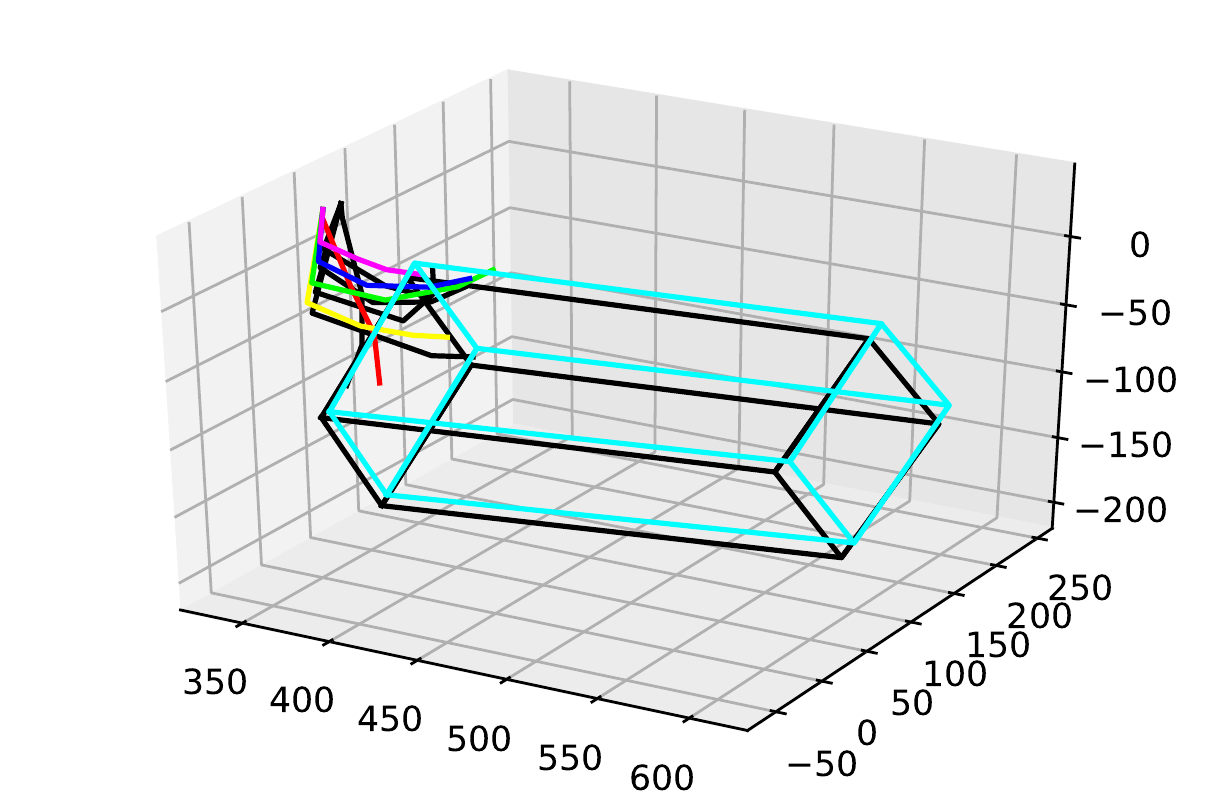} \\
  \includegraphics[scale=0.225]{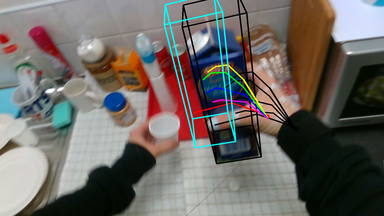} &
  \includegraphics[scale=0.21]{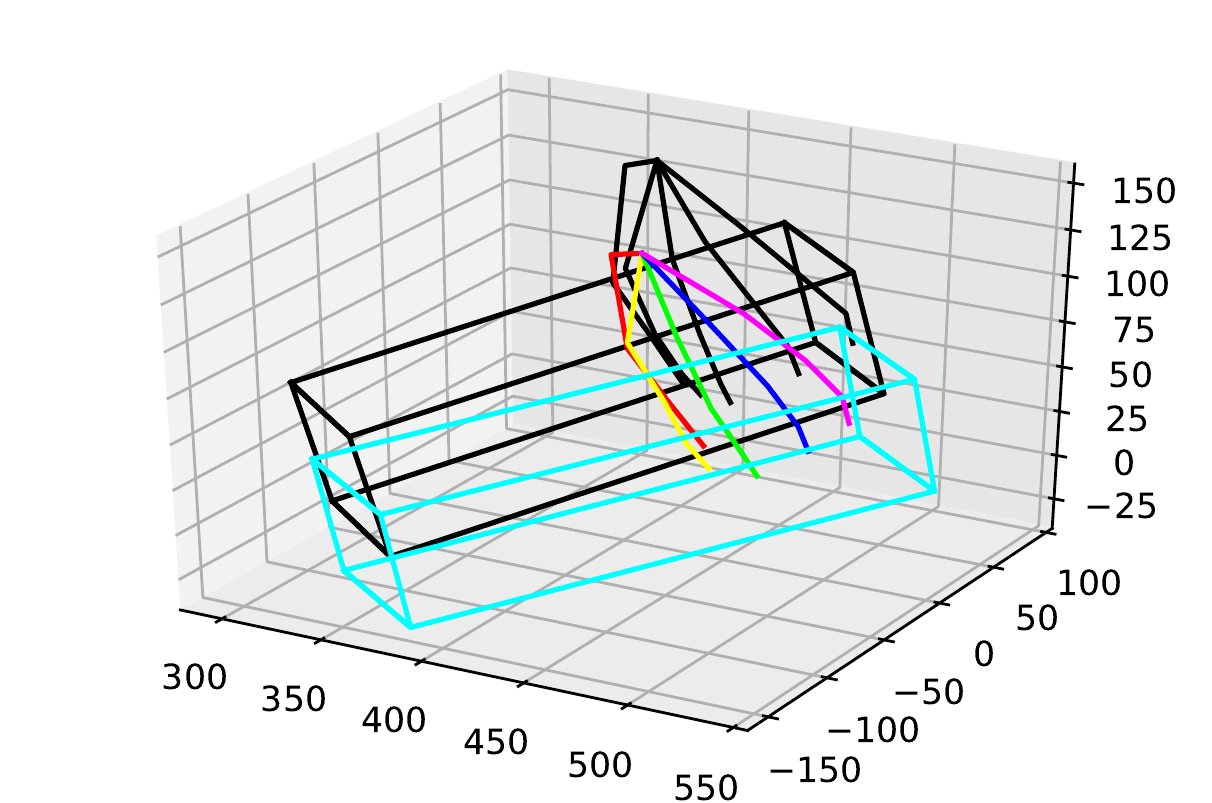} &
  \includegraphics[scale=0.225]{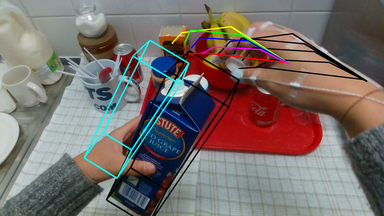} &
  \includegraphics[scale=0.21]{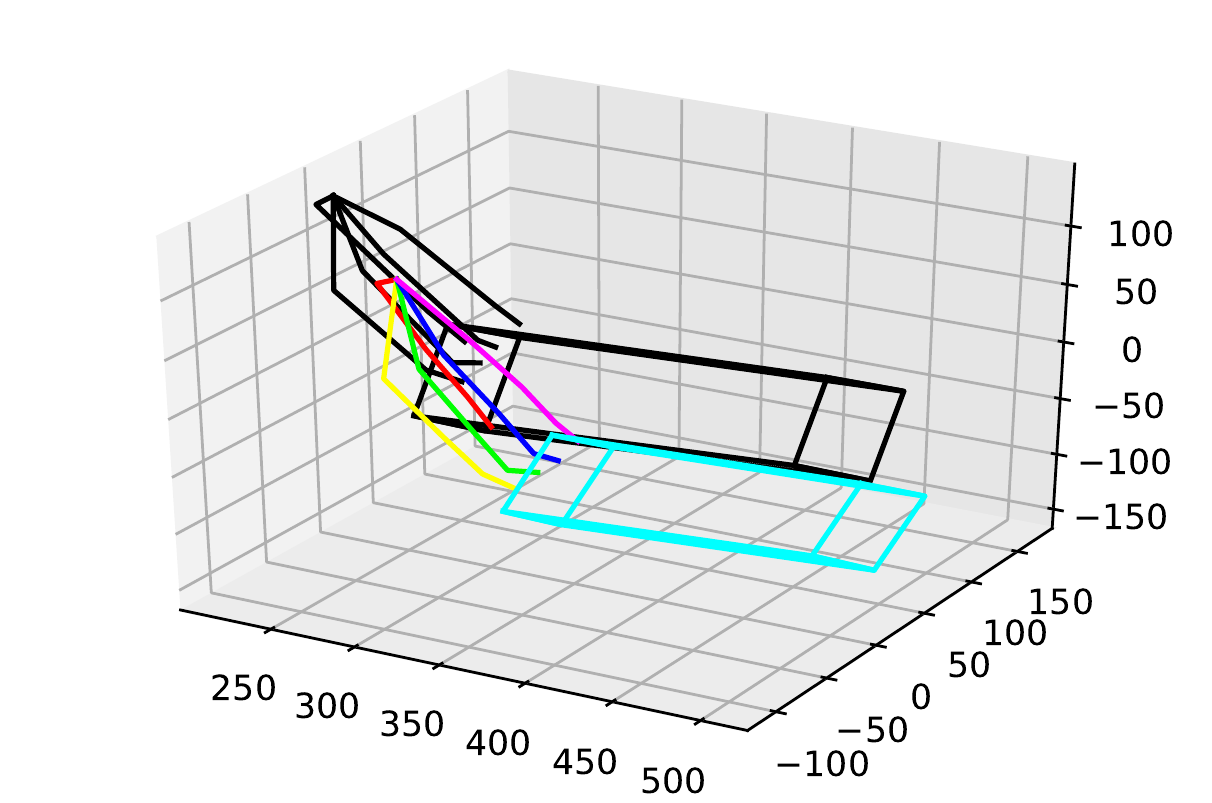} &
  \includegraphics[scale=0.225]{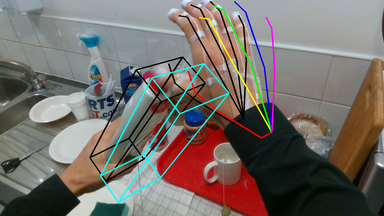} &
  \includegraphics[scale=0.21]{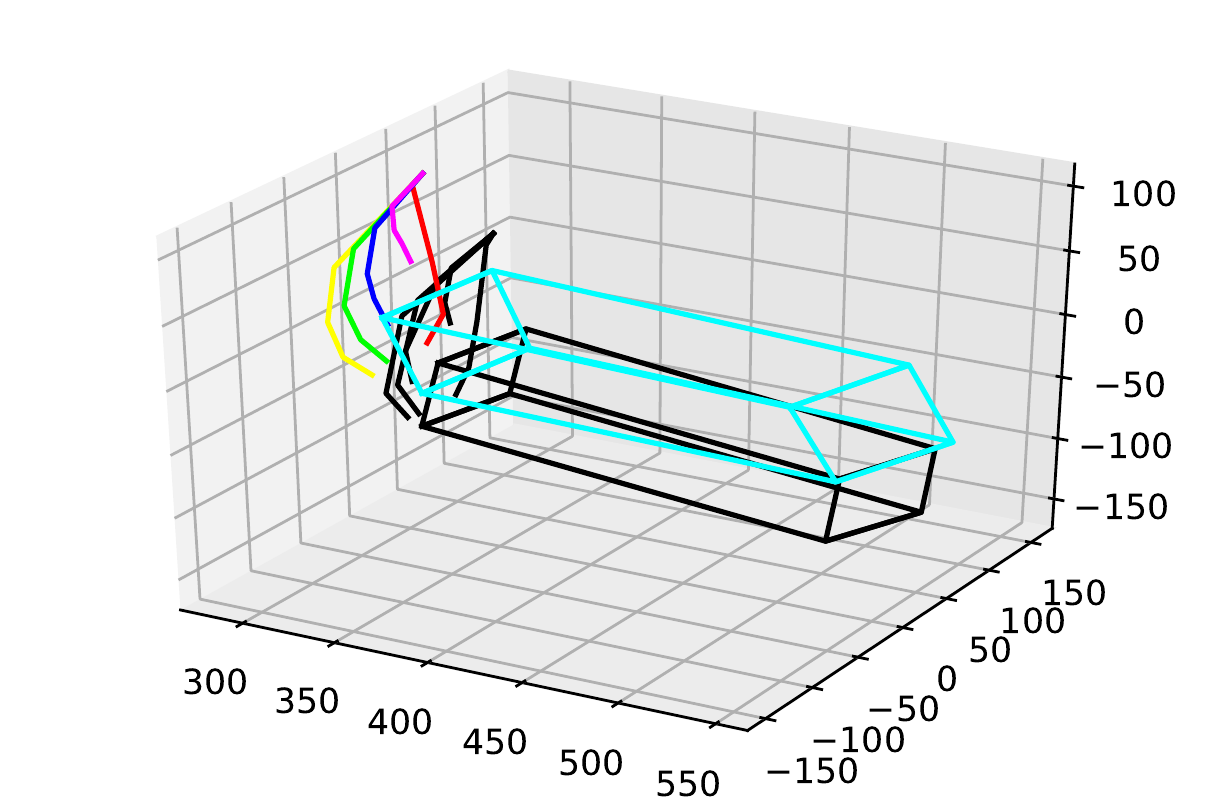}
  \end{tabular}
  \vspace{2pt}
  \caption{Qualitative 2D and 3D results of HOPE-Net on the First-Person Hand Action
    dataset. The estimated poses are shown in color,
    and the ground truth is shown in black.
    The last row includes three failure cases.
}
  \label{fig:results}
\end{figure*}

\begin{figure}[t]
  \centering
  \begin{tabular}{cc}
  \includegraphics[scale=0.55, trim=10 0 0 0]{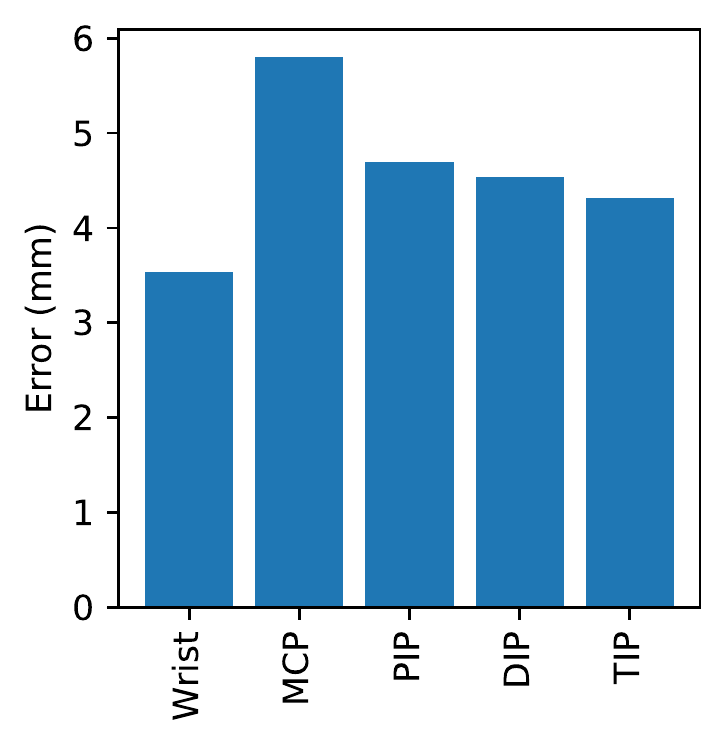} &
  \includegraphics[scale=0.55, trim=10 9 0 0]{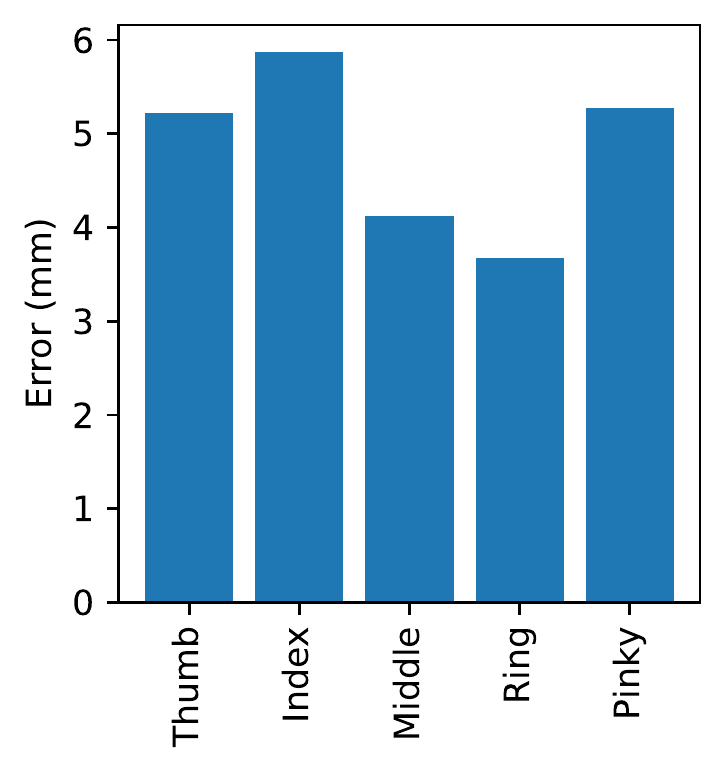} \\
  (a) & (b)
  \end{tabular}
  \vspace{1pt}
  \caption{Average 3D pose estimation errors broken out across (a) each joint of the hand and (b) each finger.
Note that MCP, PIP, and DIP denote the 3 joints located between the wrist and fingertip (TIP), in that order.
}
  \label{fig:joints}
\end{figure}

Figure~\ref{fig:results} shows selected qualitative results of our
model on the First-Person Hand Action dataset. 
Figure~\ref{fig:joints} breaks out the
error of the 2D to 3D converter for each  finger and also for
each kind of joint of the hand.

We also tested on the third-person videos of the very recent HO-3D dataset.
Although the locations of hands and objects in the images vary more in
HO-3D, we found  that HOPE-Net performs better, perhaps because of the size of the dataset.
The Area Under the
Curve (AUC) score of HOPE-Net is \(0.712\) for 2D pose 
and \(0.967\) for 3D pose estimation.
Note that hands in the evaluation set of  HO-3D  are just annotated with
the wrist (without the full hand annotation). Therefore the mentioned results
are just for wrist keypoint.

\begin{figure*}[ht]
  \centering
  \setlength{\tabcolsep}{4pt}
  \begin{tabular}{ccccc}
  \includegraphics[scale=0.35]{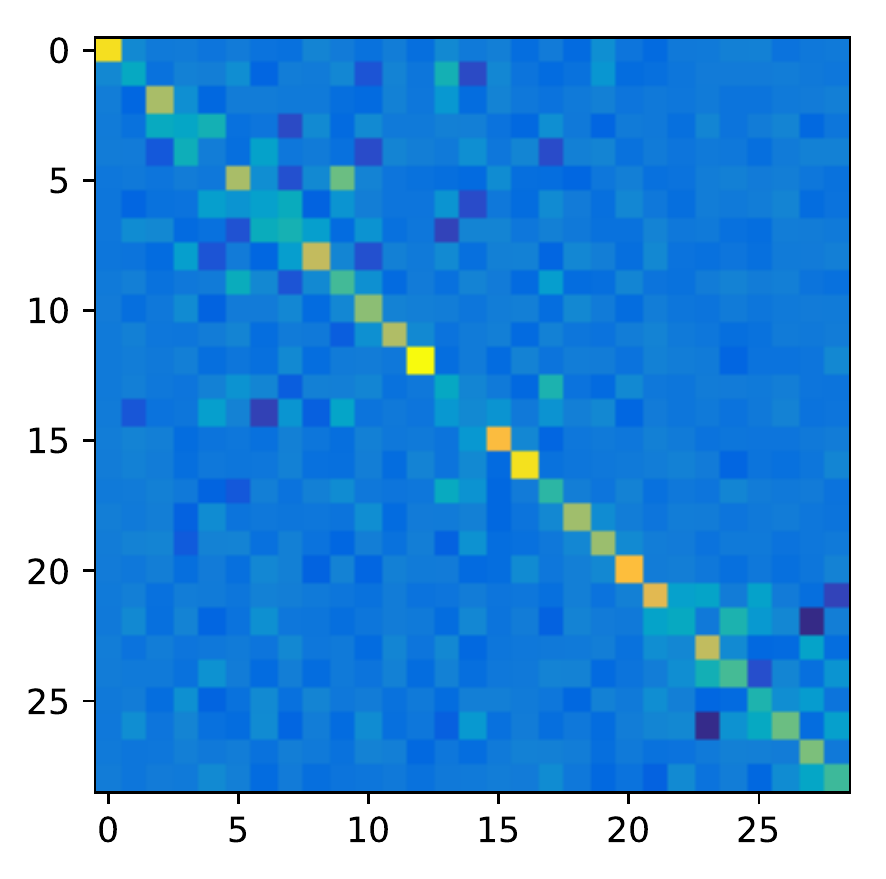} &
  \includegraphics[scale=0.35]{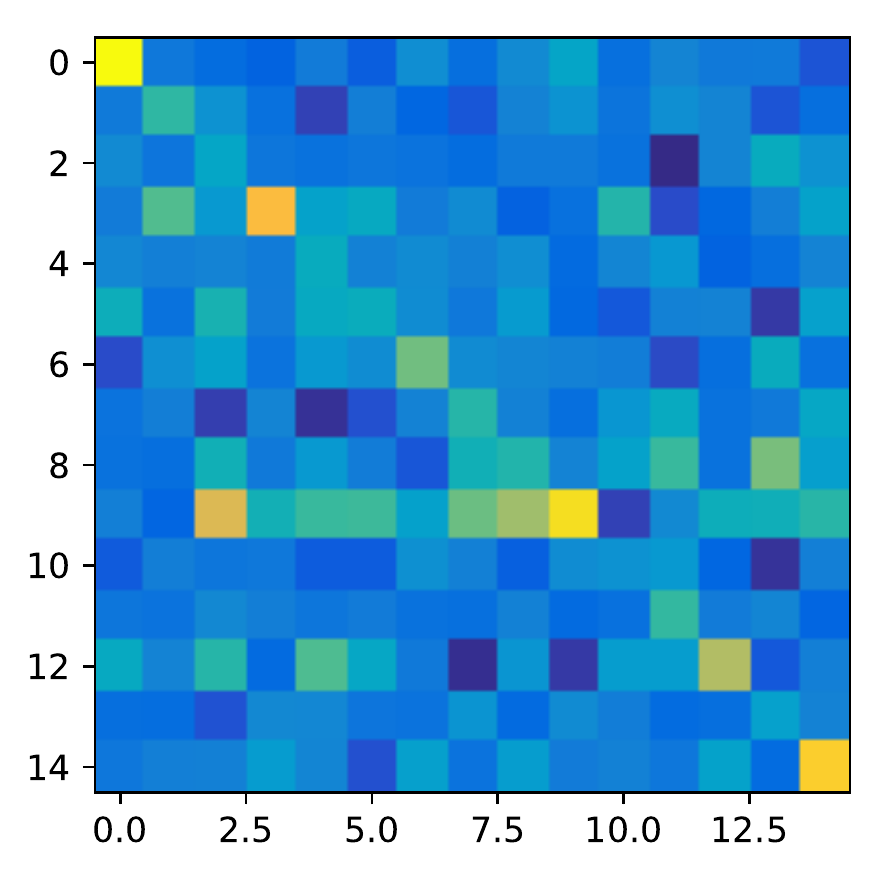} &
  \includegraphics[scale=0.35]{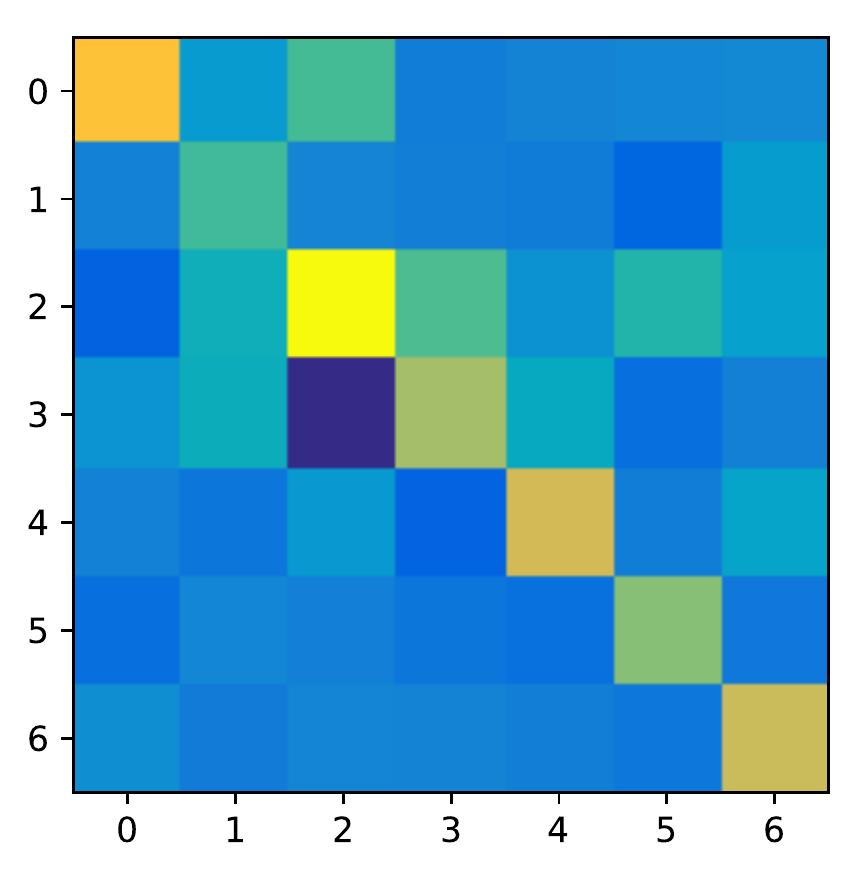} &
  \includegraphics[scale=0.35]{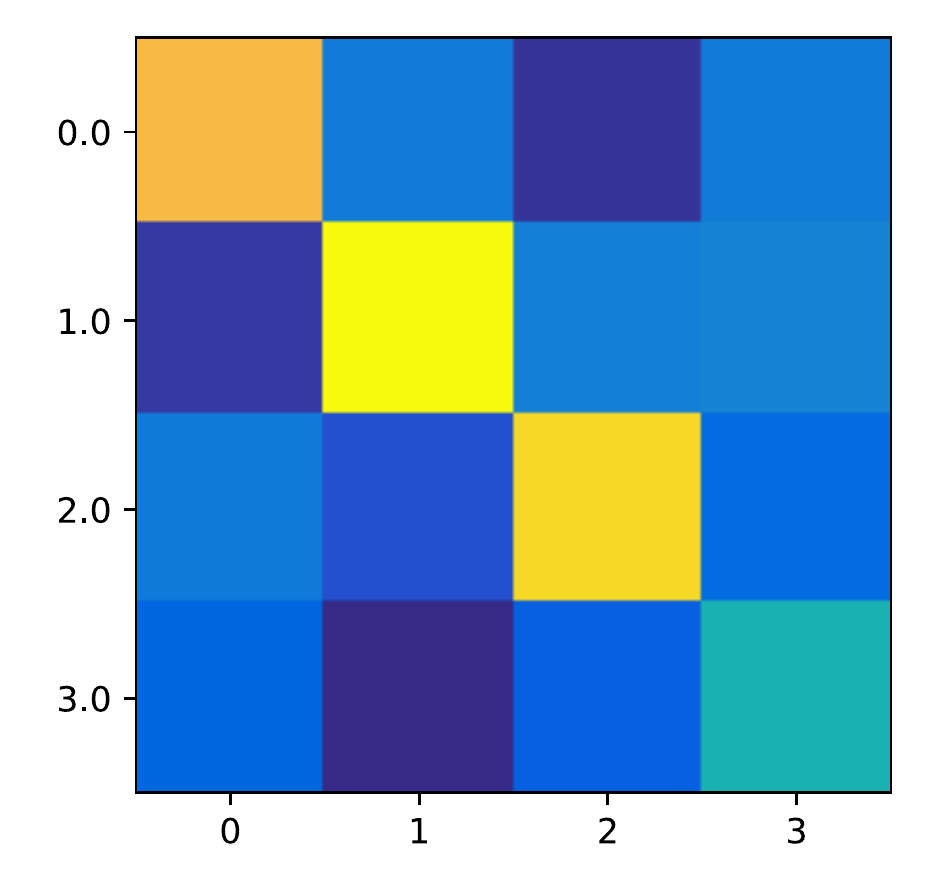} &
  \includegraphics[scale=0.35]{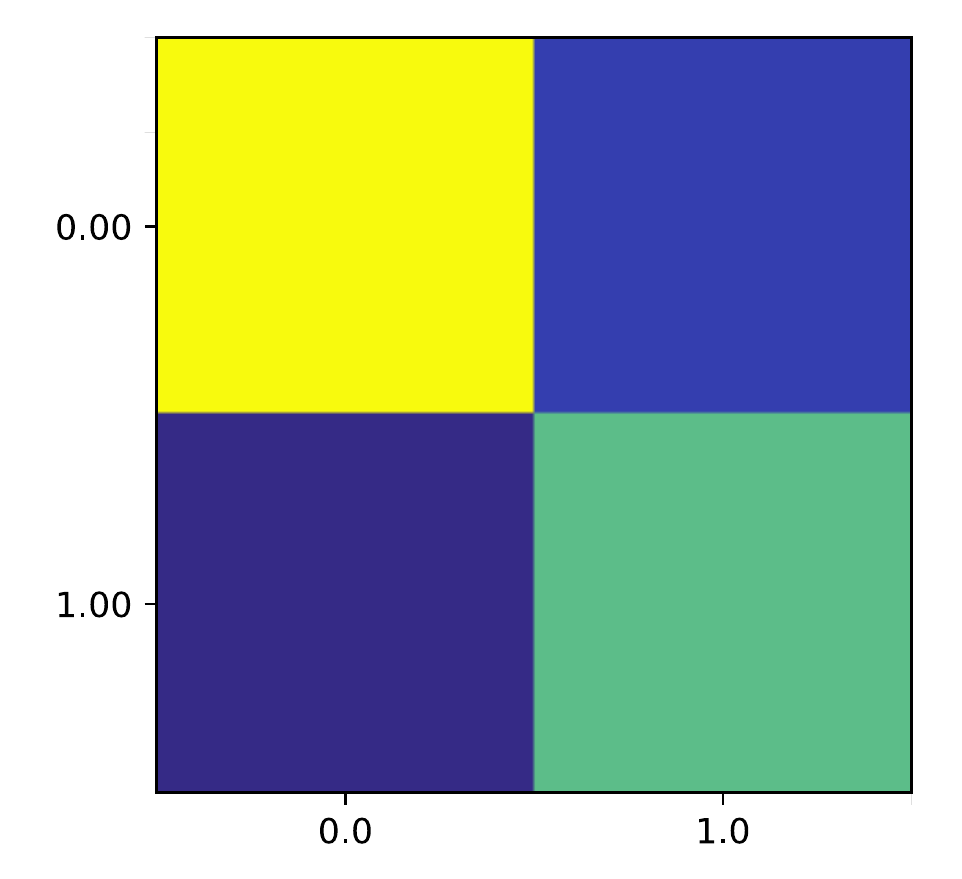} \\
  \(A_0\) & \(A_1\) & \(A_2\) & \(A_3\) & \(A_4\)
  \end{tabular}
  \vspace{1pt}
  \caption{Visualization of the learned adjacency matrices of the
    adaptive graph convolution layers. For instance, we see in the
    \(A_0\) matrix that the corners of the object bounding box (row and
    column indices 21 through 29) are highly dependent on one another, and also there is a
    relatively strong connection between fingertips.}
  \label{fig:adjacency}
\end{figure*}

\subsection{Adaptive Graph U-Net Ablation Study}
\label{sec:graphunet}
We also conducted an ablation study of our Adaptive Graph U-Net to
identify which components were important for achieving our results.
We first compare our model to other models, and then we evaluate the
influence of the adjacency matrix initialization on the adaptive Graph
U-Net performance.

To show the effectiveness of our U-Net structure, we compared it
to two different models, one with three Fully Connected Layers and one with three Graph
Convolutional Layers without pooling and unpooling. We were interested
in the importance of each of our graph convolutional models in the
3D output. Each of these models is trained to convert 2D coordinates
of the hand and object keypoints to 3D.
Table~\ref{tab:ablation} shows the results.
The
adaptive Graph U-Net performs better than the other methods by a
large margin. This large margin seems to  come from the
U-Net structure and the pooling and unpooling layers.

To understand the effect of our graph pooling layer, we compared it
with Gao \etal's~\cite{gao2019graph} gPool, and also with fixed
pooled nodes which do not break the graph into
pieces. Table~\ref{tab:pool} compares the performance of different
graph pooling methods. We see that by using a more efficient training
algorithm and also by not breaking apart the graph after pooling, our
pooling layer performs better than gPool.

\begin{table}[t]
    \centering
    \caption{Average error on 3D hand and object pose estimation given
      2D pose. The first row is a multi-layer perceptron and the
      second row  is a 3-layered graph convolution without pooling
      and unpooling. The Adaptive Graph U-Net structure
      has the best performance.}
    \vspace{5pt}
    {\small{
    \begin{tabular}{lr}\toprule
    Architecture & Average Error (mm)\\ \midrule
    Fully Connected & 185.18 \\
    Adaptive Graph Convolution & 68.93 \\
    Adaptive Graph U-Net & \textbf{6.81} \\
    \bottomrule
    \end{tabular}
    }}
    \vspace{5pt}
    \label{tab:ablation}
\end{table}

\begin{table}[t]
    \centering
    \caption{Average error on 2D to 3D hand and object pose estimation
      using different pooling methods. Our trainable
      pooling method produces the best results.}
    \vspace{5pt}
    {\small{
    \begin{tabular}{lr}\toprule
    Pooling method & Average Error (mm)\\ \midrule
    gPool~\cite{gao2019graph} & 153.28 \\
    Fixed Pooling Layers & 7.41 \\
    Trainable Pooling & \textbf{6.81} \\
    \bottomrule
    \end{tabular}
    }}
    \vspace{5pt}
    \label{tab:pool}
\end{table}

Since we are using an adaptive graph convolution, the network learns
the adjacency matrix as well. We tested the effect of different adjacency matrix
initializations on the final performance, including: 
hand skeleton and object bounding box, empty graph with and without
self-loops, complete graph, and a random connection of 
vertices. Table~\ref{tab:init} presents the results of the model initialized
with each of these matrices, showing that the identity matrix is the 
best initialization.
In other words, the model seems to learn  best when it finds the
relationship between the nodes starting with an unbiased (uninformative) initialization.

\begin{table}[t]
    \centering
    \caption{Average error in 3D pose estimation in the
      adaptive graph convolution layer. The model has the best performance when
      it is initialized with the identity matrix. ``Skeleton'' in the
      fourth row refers to an adjacency matrix that simply encodes the actual kinematic
structure of the human hand.}
    \vspace{5pt}
    {\small{
    \begin{tabular}{lr}\toprule
    Initial Adjacency Matrix & Average Error (mm)\\ \midrule
    Zeros (\(\mathbf{0}_{n \times n}\)) & 92805.02 \\
    Random Initialization & 94.42 \\
    Ones (\(\mathbf{1}_{n \times n}\)) & 63.25 \\
    Skeleton & 12.91 \\
    Identity (\(\mathbf{I}_{n \times n}\)) & \textbf{6.81} \\
    \bottomrule
    \end{tabular}
    }}
    \vspace{5pt}
    \label{tab:init}
\end{table}

The final trained adjacency matrices for the graph convolution layers
(starting from \(\mathbf{I}_{n \times n}\)) are visualized in
Figure~\ref{fig:adjacency}.
We see that the model has found relationships between  nodes which are not
connected in the hand skeleton model. For example, it found a relationship between
node 6 (index finger's PIP) and node 4 (thumb's TIP), which are not connected in the
hand skeleton model.

\subsection{Runtime}
As  mentioned earlier, HOPE-Net consists of a lightweight
feature extractor (ResNet10) and two graph convolutional neural networks which
are more than ten times faster than the shallowest image convolutional
neural network. The core inference of the model can be run in
real-time on an Nvidia Titan Xp. On such a GPU, the entire 2D and 3D
inference of a single frame requires just 0.005 seconds.

\section{Conclusion and Future Work}
In this paper, we introduced a model for hand-object 2D and 3D pose
estimation from a single image using an image encoder followed by a
cascade of two graph convolutional neural networks.  Our approach
beats the state-of-the-art, while also running in real-time.

Nevertheless, there are limitations of our approach.  When trained on
the FPHA and HO-3D datasets, our model is well-suited for objects that
are of similar size or shape to those seen in the dataset during
training, but might not generalize well to all categories of object
shapes. For example, objects with a non-convex geometry lacking a
tight 3D bounding box would be a challenge for our technique. For
real-world applications, a larger dataset including a greater variety
of shapes and environments would help to improve the estimation
accuracies.

Future work could include incorporating temporal information into our
graph-based model, both to improve pose estimation results and as
step towards action detection. Graph classification methods can be integrated
into the proposed framework to infer categorical semantic information
for applications such as detecting sign language or gesture
understanding. Also, in addition to hand pose estimation, the
Adaptive Graph U-Net introduced in this work can be applied to a
variety of other problems such as graph completion,
protein classification, mesh classification, and body pose estimation.

\section*{Acknowledgment}
The work in this paper was supported in part by  the
National Science Foundation (CAREER IIS-1253549), and by
the IU Office of the Vice Provost for Research, the College of Arts and Sciences, and the Luddy School of Informatics, Computing, and Engineering through the Emerging Areas of Research Project ``Learning: Brains, Machines, and Children.''

{\small
\bibliographystyle{ieee_fullname}
\bibliography{main}

\begin{thebibliography}{10}\itemsep=-1pt

\bibitem{cai2019exploiting}
Yujun Cai, Liuhao Ge, Jun Liu, Jianfei Cai, Tat-Jen Cham, Junsong Yuan, and
  Nadia~Magnenat Thalmann.
\newblock Exploiting spatial-temporal relationships for 3d pose estimation via
  graph convolutional networks.
\newblock In {\em Proceedings of the IEEE International Conference on Computer
  Vision}, pages 2272--2281, 2019.

\bibitem{chang2015shapenet}
Angel~X Chang, Thomas Funkhouser, Leonidas Guibas, Pat Hanrahan, Qixing Huang,
  Zimo Li, Silvio Savarese, Manolis Savva, Shuran Song, Hao Su, et~al.
\newblock Shapenet: An information-rich 3d model repository.
\newblock {\em arXiv preprint arXiv:1512.03012}, 2015.

\bibitem{choi2017robust}
Chiho Choi, Sang Ho~Yoon, Chin-Ning Chen, and Karthik Ramani.
\newblock Robust hand pose estimation during the interaction with an unknown
  object.
\newblock In {\em IEEE International Conference on Computer Vision (ICCV)},
  pages 3123--3132, 2017.

\bibitem{survey}
Bardia Doosti.
\newblock Hand pose estimation: {A} survey.
\newblock {\em CoRR}, abs/1903.01013, 2019.

\bibitem{gao2019graph}
Hongyang Gao and Shuiwang Ji.
\newblock Graph u-nets.
\newblock In {\em International Conference on Learning Representations (ICLR)},
  2019.

\bibitem{FirstPersonActionDataset}
Guillermo Garcia-Hernando, Shanxin Yuan, Seungryul Baek, and Tae-Kyun Kim.
\newblock First-person hand action benchmark with rgb-d videos and 3d hand pose
  annotations.
\newblock In {\em IEEE Conference on Computer Vision and Pattern Recognition
  (CVPR)}, 2018.

\bibitem{ge20193d}
Liuhao Ge, Zhou Ren, Yuncheng Li, Zehao Xue, Yingying Wang, Jianfei Cai, and
  Junsong Yuan.
\newblock 3d hand shape and pose estimation from a single rgb image.
\newblock In {\em IEEE Conference on Computer Vision and Pattern Recognition
  (CVPR)}, pages 10833--10842, 2019.

\bibitem{hampali2019honnotate}
Shreyas Hampali, Mahdi Rad, Markus Oberweger, and Vincent Lepetit.
\newblock Honnotate: A method for 3d annotation of hand and objects poses,
  2019.

\bibitem{hanocka2019meshcnn}
Rana Hanocka, Amir Hertz, Noa Fish, Raja Giryes, Shachar Fleishman, and Daniel
  Cohen-Or.
\newblock Meshcnn: a network with an edge.
\newblock {\em ACM Transactions on Graphics (TOG)}, 38(4):90, 2019.

\bibitem{hasson19_obman}
Yana Hasson, G{\"u}l Varol, Dimitrios Tzionas, Igor Kalevatykh, Michael~J.
  Black, Ivan Laptev, and Cordelia Schmid.
\newblock Learning joint reconstruction of hands and manipulated objects.
\newblock In {\em IEEE Conference on Computer Vision and Pattern Recognition
  (CVPR)}, 2019.

\bibitem{resnet}
Kaiming He, Xiangyu Zhang, Shaoqing Ren, and Jian Sun.
\newblock Deep residual learning for image recognition.
\newblock In {\em IEEE Conference on Computer Vision and Pattern Recognition
  (CVPR)}, pages 770--778, 2016.

\bibitem{ioffe2015batch}
Sergey Ioffe and Christian Szegedy.
\newblock Batch normalization: Accelerating deep network training by reducing
  internal covariate shift.
\newblock In {\em International Conference on Machine Learning (ICML)}, 2015.

\bibitem{kipf2017semi}
Thomas~N Kipf and Max Welling.
\newblock Semi-supervised classification with graph convolutional networks.
\newblock In {\em International Conference on Learning Representations (ICLR)},
  2017.

\bibitem{kolotouros2019convolutional}
Nikos Kolotouros, Georgios Pavlakos, and Kostas Daniilidis.
\newblock Convolutional mesh regression for single-image human shape
  reconstruction.
\newblock In {\em IEEE Conference on Computer Vision and Pattern Recognition
  (CVPR)}, pages 4501--4510, 2019.

\bibitem{Kolotouros_2019_CVPR}
Nikos Kolotouros, Georgios Pavlakos, and Kostas Daniilidis.
\newblock Convolutional mesh regression for single-image human shape
  reconstruction.
\newblock In {\em IEEE Conference on Computer Vision and Pattern Recognition
  (CVPR)}, June 2019.

\bibitem{krizhevsky2012imagenet}
Alex Krizhevsky, Ilya Sutskever, and Geoffrey~E Hinton.
\newblock Imagenet classification with deep convolutional neural networks.
\newblock In {\em Advances in Neural Information Processing Systems (NeurIPS)},
  pages 1097--1105, 2012.

\bibitem{li2019skeleton}
Maosen Li, Siheng Chen, Xu Chen, Ya Zhang, Yanfeng Wang, and Qi Tian.
\newblock Actional-structural graph convolutional networks for skeleton-based
  action recognition.
\newblock In {\em IEEE Conference on Computer Vision and Pattern Recognition
  (CVPR)}, June 2019.

\bibitem{nair2010rectified}
Vinod Nair and Geoffrey~E Hinton.
\newblock Rectified linear units improve restricted boltzmann machines.
\newblock In {\em International Conference on Machine Learning (ICML)}, pages
  807--814, 2010.

\bibitem{oberweger2019generalized}
Markus Oberweger, Paul Wohlhart, and Vincent Lepetit.
\newblock Generalized feedback loop for joint hand-object pose estimation.
\newblock {\em IEEE Transactions on Pattern Analysis and Machine Intelligence},
  2019.

\bibitem{oikonomidis2011full}
Iason Oikonomidis, Nikolaos Kyriazis, and Antonis~A Argyros.
\newblock Full dof tracking of a hand interacting with an object by modeling
  occlusions and physical constraints.
\newblock In {\em 2011 International Conference on Computer Vision}, pages
  2088--2095. IEEE, 2011.

\bibitem{plastira3d}
Paschalis Panteleris, Nikolaos Kyriazis, and Antonis~A Argyros.
\newblock 3d tracking of human hands in interaction with unknown objects.
\newblock In {\em British Machine Vision Conference (BMVC)}, pages 123--1,
  2015.

\bibitem{ranjan2018generating}
Anurag Ranjan, Timo Bolkart, Soubhik Sanyal, and Michael~J Black.
\newblock Generating 3d faces using convolutional mesh autoencoders.
\newblock In {\em European Conference on Computer Vision (ECCV)}, pages
  704--720, 2018.

\bibitem{ronneberger2015unet}
Olaf Ronneberger, Philipp Fischer, and Thomas Brox.
\newblock U-net: Convolutional networks for biomedical image segmentation.
\newblock In {\em International Conference on Medical Image Computing and
  Computer-assisted Intervention}, pages 234--241. Springer, 2015.

\bibitem{shi2019skeleton}
Lei Shi, Yifan Zhang, Jian Cheng, and Hanqing Lu.
\newblock Skeleton-based action recognition with directed graph neural
  networks.
\newblock In {\em IEEE Conference on Computer Vision and Pattern Recognition
  (CVPR)}, June 2019.

\bibitem{shi2019adaptive}
Lei Shi, Yifan Zhang, Jian Cheng, and Hanqing Lu.
\newblock Two-stream adaptive graph convolutional networks for skeleton-based
  action recognition.
\newblock In {\em IEEE Conference on Computer Vision and Pattern Recognition
  (CVPR)}, June 2019.

\bibitem{tekin2019unified}
Bugra Tekin, Federica Bogo, and Marc Pollefeys.
\newblock H+o: Unified egocentric recognition of 3d hand-object poses and
  interactions.
\newblock In {\em IEEE Conference on Computer Vision and Pattern Recognition
  (CVPR)}, June 2019.

\bibitem{tekin2018real}
Bugra Tekin, Sudipta~N Sinha, and Pascal Fua.
\newblock Real-time seamless single shot 6d object pose prediction.
\newblock In {\em IEEE Conference on Computer Vision and Pattern Recognition
  (CVPR)}, pages 292--301, 2018.

\bibitem{wu2018group}
Yuxin Wu and Kaiming He.
\newblock Group normalization.
\newblock In {\em European Conference on Computer Vision (ECCV)}, pages 3--19,
  2018.

\bibitem{yan2018spatial}
Sijie Yan, Yuanjun Xiong, and Dahua Lin.
\newblock Spatial temporal graph convolutional networks for skeleton-based
  action recognition.
\newblock In {\em Thirty-Second AAAI Conference on Artificial Intelligence},
  2018.

\bibitem{yuan2018depth}
Shanxin Yuan, Guillermo Garcia-Hernando, Bj{\"o}rn Stenger, Gyeongsik Moon, Ju
  Yong~Chang, Kyoung Mu~Lee, Pavlo Molchanov, Jan Kautz, Sina Honari, Liuhao
  Ge, et~al.
\newblock Depth-based 3d hand pose estimation: From current achievements to
  future goals.
\newblock In {\em IEEE Conference on Computer Vision and Pattern Recognition
  (CVPR)}, pages 2636--2645, 2018.

\bibitem{zhao2019semantic}
Long Zhao, Xi Peng, Yu Tian, Mubbasir Kapadia, and Dimitris~N Metaxas.
\newblock Semantic graph convolutional networks for 3d human pose regression.
\newblock In {\em IEEE Conference on Computer Vision and Pattern Recognition
  (CVPR)}, pages 3425--3435, 2019.

\end{thebibliography}
}

\end{document}